\documentclass[letterpaper,10pt,journal,twoside]{IEEEtran}

\IEEEoverridecommandlockouts                              

\usepackage{xr}

\newcommand{\omitted}[1]{}

\title{
\fontsize{22}{22} \selectfont 
Adaptive Legged Locomotion via \\ Online Learning for Model Predictive Control
}

\author{
\thanks{Manuscript received June 15, 2025; Revised October 5, 2025; Accepted November 25, 2025. This paper was recommended for publication by Editor Abderrahmane Kheddar upon evaluation of the Associate Editor and Reviewers’ comments. (Corresponding author: Hongyu Zhou.)}
Hongyu Zhou$^{\dagger,1}$ \;\; Xiaoyu Zhang$^{\dagger,2}$  \;\; Vasileios Tzoumas$^1$
\thanks{$^{\dagger}$Equal contribution}
\thanks{$^{1}$Department of Aerospace Engineering, University of Michigan, Ann Arbor, MI 48109 USA;  {\tt\footnotesize \{zhouhy, vtzoumas\}@umich.edu}}
\thanks{$^{2}$Institute for Robotics and Intelligent Machines, Georgia Institute of Technology, Atlanta, GA 30332, USA; {\tt\footnotesize xzhang636@gatech.edu}}
\thanks{This work was partially supported by NSF CAREER No. 2337412 and partially by Rackham Predoctoral Fellowship of the University of Michigan.}
\thanks{Digital Object Identifier (DOI): see top of this page.}
}

\usepackage{cite}

\usepackage{comment}
\usepackage{siunitx}
\usepackage{relsize}
\usepackage{ifthen}
\usepackage[colorinlistoftodos]{todonotes}
\usepackage[vlined,ruled,linesnumbered]{algorithm2e}
\usepackage{graphics} 
\usepackage{rotating}
\usepackage{color}
\usepackage{enumerate}
\usepackage[T1]{fontenc}
\usepackage{psfrag}
\usepackage{epsfig} 
\usepackage{booktabs}
\usepackage{graphicx}
\usepackage{url}
\usepackage{multirow}
\usepackage{array}
\usepackage{latexsym}
\usepackage{amsfonts}
\usepackage{amsmath}
\usepackage{amssymb}
\usepackage{xstring}
\usepackage{algorithmic}
\usepackage{multirow}
\usepackage{xcolor}
\usepackage{prettyref}
\usepackage{flexisym}
\usepackage{bigdelim}
\usepackage{breqn} 
\usepackage{listings}

\usepackage{xspace}
\usepackage{bm}
\usepackage{soul}
\graphicspath{{./figures/}}
\usepackage{tikz}
\usetikzlibrary{matrix,calc}
\usepackage{lipsum}
\usepackage{mdwlist}

\let\itemize\undefined
\makecompactlist{itemize}{stditemize}
\usepackage{enumitem}
\usepackage{caption}
\usepackage{epstopdf}
\usepackage{subfigure}

\usepackage{amsthm}
\usepackage{mathtools}
\usepackage{gensymb}

\makeatletter
\let\NAT@parse\undefined
\makeatother
\usepackage{hyperref}
\usepackage{cleveref}

\hypersetup{%
  colorlinks=true,
  linkcolor=blue,
  filecolor=magenta,      
  urlcolor=cyan,
  citecolor=red,
  linkbordercolor={0 0 1}
}

\newtheorem{theorem}{Theorem}
\newtheorem{problem}{Problem}

\newtheorem{definition}{Definition}
\newtheorem{proposition}{Proposition}

\newcommand{\bdmath}{\begin{dmath}}
\newcommand{\edmath}{\end{dmath}}
\newcommand{\beq}{\begin{equation}}
\newcommand{\eeq}{\end{equation}}
\newcommand{\bdm}{\begin{displaymath}}
\newcommand{\edm}{\end{displaymath}}
\newcommand{\bea}{\begin{eqnarray}}
\newcommand{\eea}{\end{eqnarray}}
\newcommand{\beal}{\beq \begin{array}{lll}}
\newcommand{\eeal}{\end{array} \eeq}
\newcommand{\beas}{\begin{eqnarray*}}
\newcommand{\eeas}{\end{eqnarray*}}
\newcommand{\ba}{\begin{array}}
\newcommand{\ea}{\end{array}}
\newcommand{\bit}{\begin{itemize}}
\newcommand{\eit}{\end{itemize}}
\newcommand{\ben}{\begin{enumerate}}
\newcommand{\een}{\end{enumerate}}

\newcommand{\calD}{{\cal D}}

\newcommand{\calF}{{\cal F}}

\newcommand{\calH}{{\cal H}}

\newcommand{\calO}{{\cal O}}

\newcommand{\calU}{{\cal U}}

\definecolor{myblue}{RGB}{65 105 225}

\newcommand{\hide}[1]{}

\newcommand{\hiddenText}{{\color{gray} hidden text.}}
\newcommand{\hideWithText}[1]{\hiddenText}

\newcommand{\diag}[1]{\mathrm{diag}\left(#1\right)}

\newcommand{\scenario}[1]{{\fontsize{9}{8.7}\selectfont\sf#1}\xspace}
\newcommand{\scenariot}[1]{{\fontsize{8}{8}\selectfont\sf#1}\xspace}
\newcommand{\scenariof}[1]{{\fontsize{7}{7}\selectfont\sf#1}\xspace}

\newcommand{\ie}{\emph{i.e.},\xspace}
\newcommand{\eg}{\emph{e.g.},\xspace}

\newcommand{\blue}[1]{{\color{blue}#1}}

\newcommand{\myParagraph}[1]{{\bf #1.}\xspace}

\newcommand{\OCO}{\scenario{{OCO}}}
\newcommand{\RKHS}{\scenario{{RKHS}}}

\newcommand{\OGD}{\scenario{{OGD}}}

\newcommand{\MPC}{\scenario{{MPC}}}

\newcommand{\NMPC}{\scenario{{Nominal MPC}}}
\newcommand{\NMPCf}{\scenariof{{Nominal MPC}}}

\newcommand{\LMPC}{\scenario{{L1-MPC}}}
\newcommand{\LMPCf}{\scenariof{{L1-MPC}}}
\newcommand{\SReg}{\operatorname{Regret}_T^S}
\newcommand{\DReg}{\operatorname{Regret}_T^D}

\PassOptionsToPackage{end}{algorithmic}

\begin{document}

\makeatletter

\g@addto@macro\@maketitle{
\setcounter{figure}{0}
\centering
\includegraphics[width=0.24\textwidth]{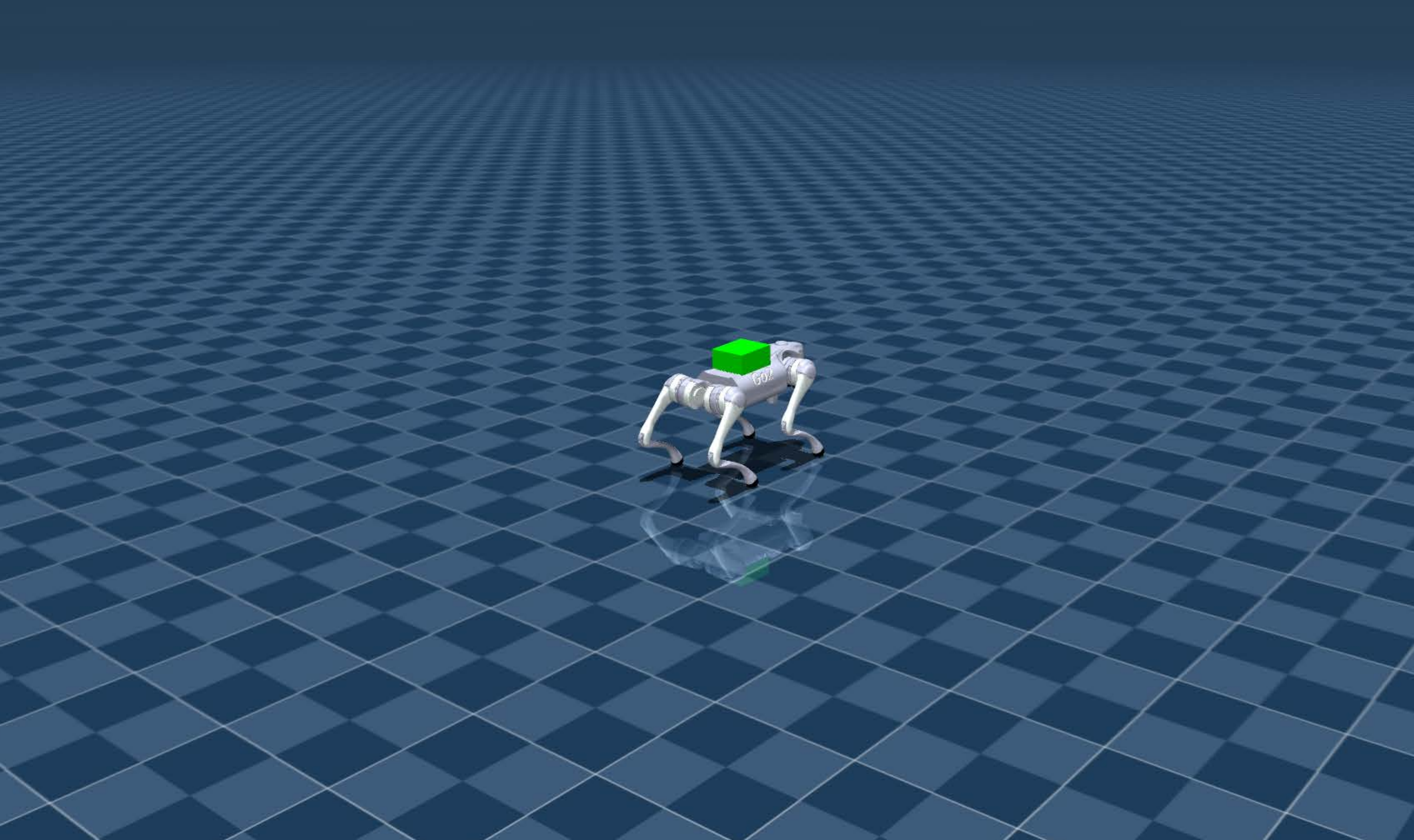}
\includegraphics[width=0.24\textwidth]{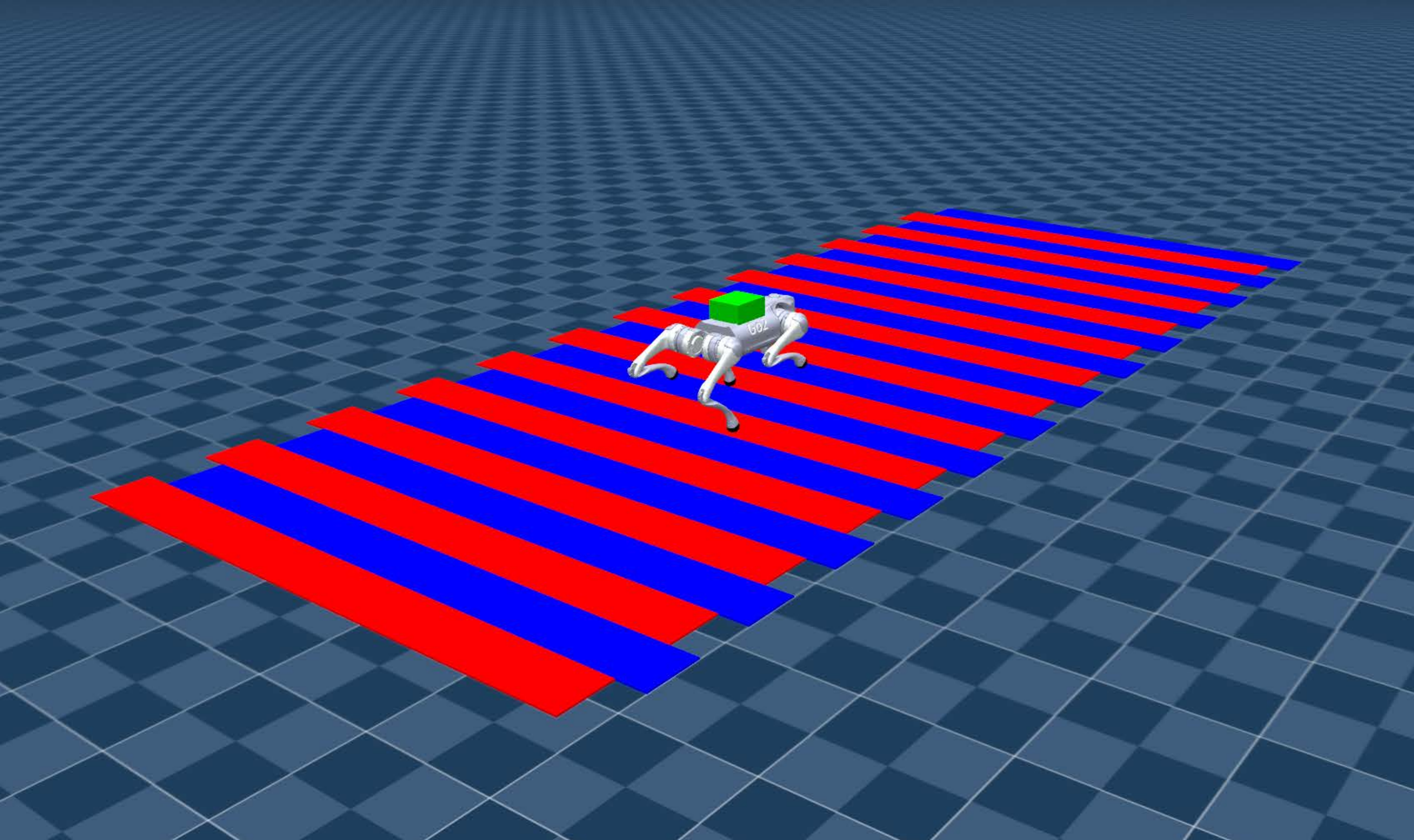}
\includegraphics[width=0.24\textwidth]{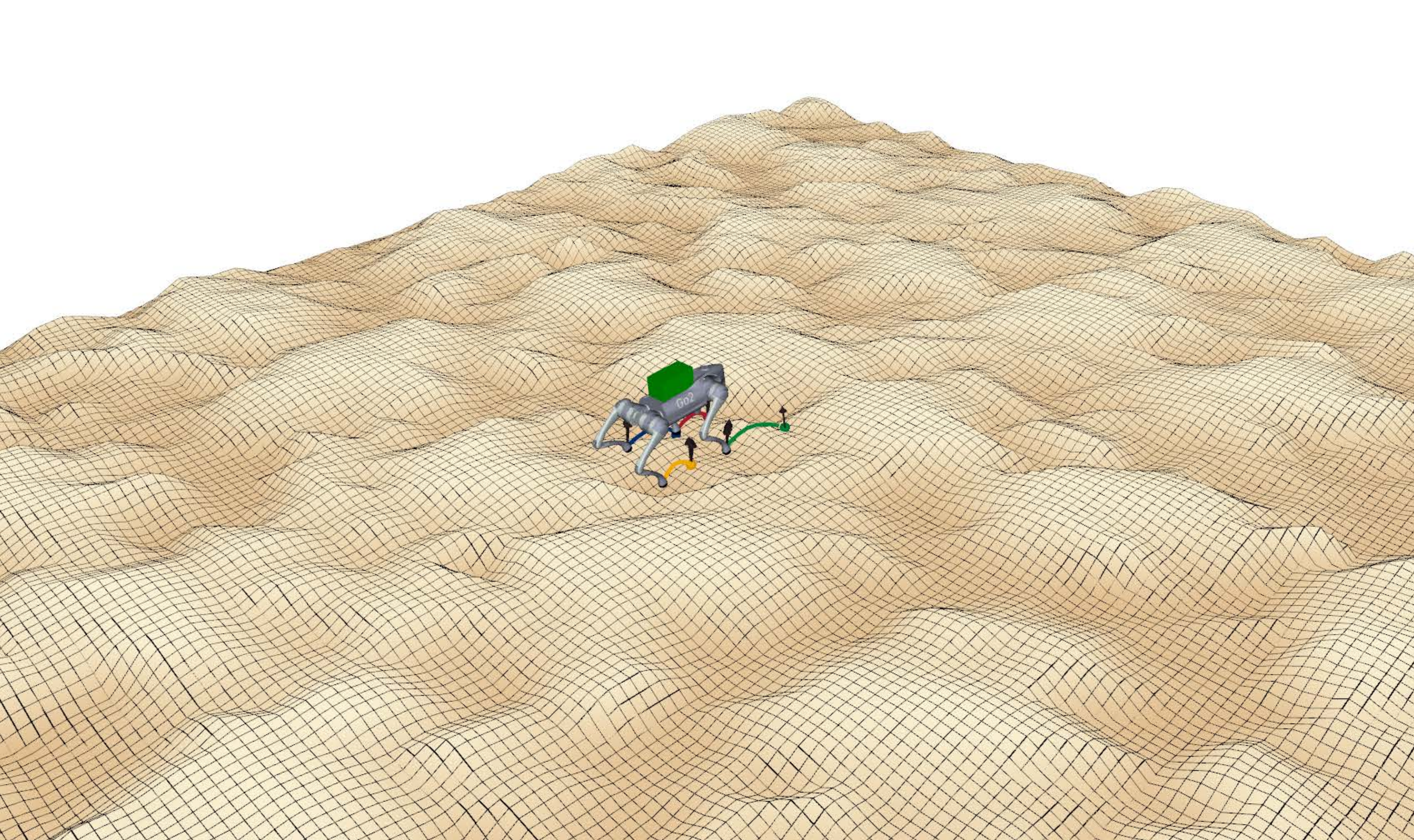}
\includegraphics[width=0.24\textwidth]{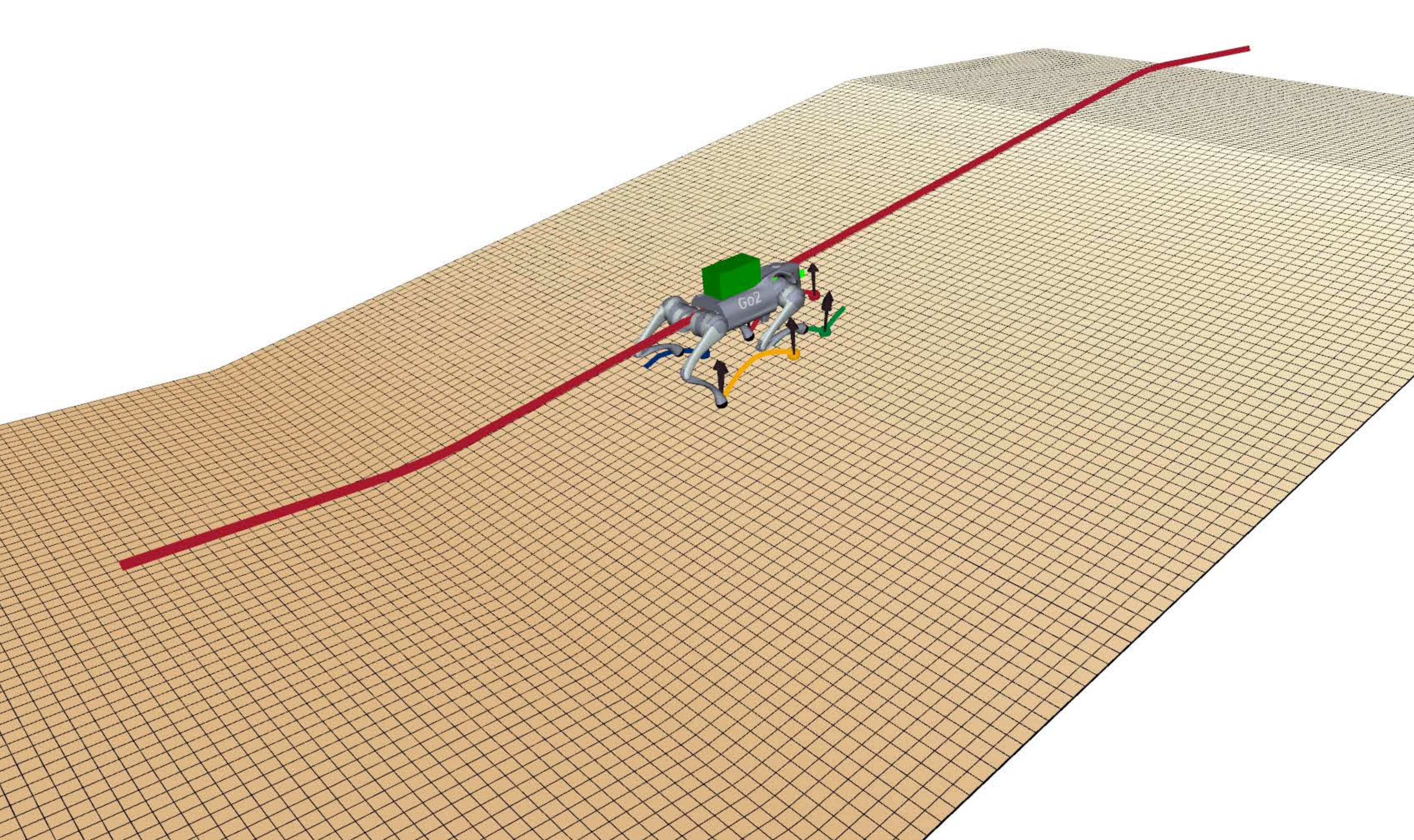}
\captionsetup{font=small}
\captionof{figure}{
\textbf{Adaptive Legged Locomotion via Online Learning and Model Predictive Control: Tested Scenarios of Reference Trajectory Tracking subject to Various Unknown Disturbances.} 
In this paper, we focus on adaptive legged locomotion via online learning and model predictive control, where the quadruped aims to achieve accurate reference tracking control despite unknown dynamics and external disturbances (\ie residual dynamics). 
For example, the reference tracking performance of the quadruped can be challenged by \textit{various terrain}~(first: flat terrain with fixed friction coefficient \& second: flat terrain with changing friction coefficients (indicate by colors) \& third: rough terrain \& fourth: slope terrain with $20\degree$ inclination) and \textit{payloads with unknown weights}~(all). We provide a control algorithm that demonstrates adaptive legged locomotion by learning such unknown dynamics/disturbances online based on the data collected at runtime, and using the learned model of residual dynamics for predictive control. 
\label{fig:sim-exp}}
\vspace{-3mm}
}

\makeatother
\maketitle


\begin{abstract}

We provide an algorithm for adaptive legged locomotion via online learning and model predictive control. 
The algorithm is composed of two interacting modules: model predictive control (MPC) and online learning of residual dynamics.
The residual dynamics can represent modeling errors and external disturbances.
We are motivated by the future of autonomy where quadrupeds will autonomously perform complex tasks despite real-world unknown uncertainty, such as unknown payload and uneven terrains.
The algorithm uses random Fourier features to approximate the residual dynamics in reproducing kernel Hilbert spaces. Then, it employs MPC based on the current learned model of the residual dynamics.
The model is updated online in a self-supervised manner using least squares based on the data collected while controlling the quadruped.
The algorithm enjoys sublinear \textit{dynamic regret}, defined as the suboptimality against an optimal clairvoyant controller that knows how the residual dynamics.
We validate our algorithm in Gazebo and MuJoCo simulations, where the quadruped aims to track reference trajectories. 
The Gazebo simulations include constant unknown external forces up to $12\boldsymbol{g}$, where $\boldsymbol{g}$ is the gravity vector, in flat terrain, slope terrain with $20\degree$ inclination, and rough terrain with $0.25m$ height variation. 
The MuJoCo simulations include time-varying unknown disturbances with payload up to $8~kg$ and time-varying ground friction coefficients in flat terrain.
The code is open-sourced at \url{https://github.com/UM-iRaL/Adaptive-Legged-Locomotion}.
\end{abstract}


\section{Introduction}\label{sec:Intro}
Legged robots promise to automate essential tasks such as search and rescue, payload delivery, and industrial inspection~\cite{seneviratne2018smart}.
Successfully accomplishing these tasks necessitates accurate and efficient tracking.
However, achieving both accuracy and efficiency in legged locomotion is challenging due to uncertainties arising from the robot's imperfect model and environmental disturbances.
For example, quadrupeds need to (i) pick up and transport packages of unknown weight and (ii) navigate diverse terrains with varying friction and elevation.

State-of-the-art methods for legged control under uncertainties~(\ie residual dynamics) typically rely on reinforcement learning methods~\cite{tan2018sim,hwangbo2019learning,lee2020learning,margolis2022walktheseways,gangapurwala2022rloc,kumar2021rma,ji2022concurrent,nahrendra2023dreamwaq,zhong2025bridging} or robust and adaptive control methods~\cite{pandala2022robust,xu2023robust,minniti2021adaptive,sun2021online,sombolestan2024adaptive,elobaid2025adaptive}.
The reinforcement learning methods require offline training and high-fidelity simulators, which {can be} costly and time-consuming.
The robust control methods can be conservative due to the assumption of {worst-case} disturbance realization and can be computationally expensive for real-time control of legged locomotion.
The adaptive control methods assume parametric uncertainty additive to the known
system dynamics and update these coefficients online to enhance the robustness against disturbances, but they often assume the uncertainty to be either vector-valued or a linear function of the state.

\begin{figure}[t]
    \centering
    \includegraphics[width=0.49\textwidth]{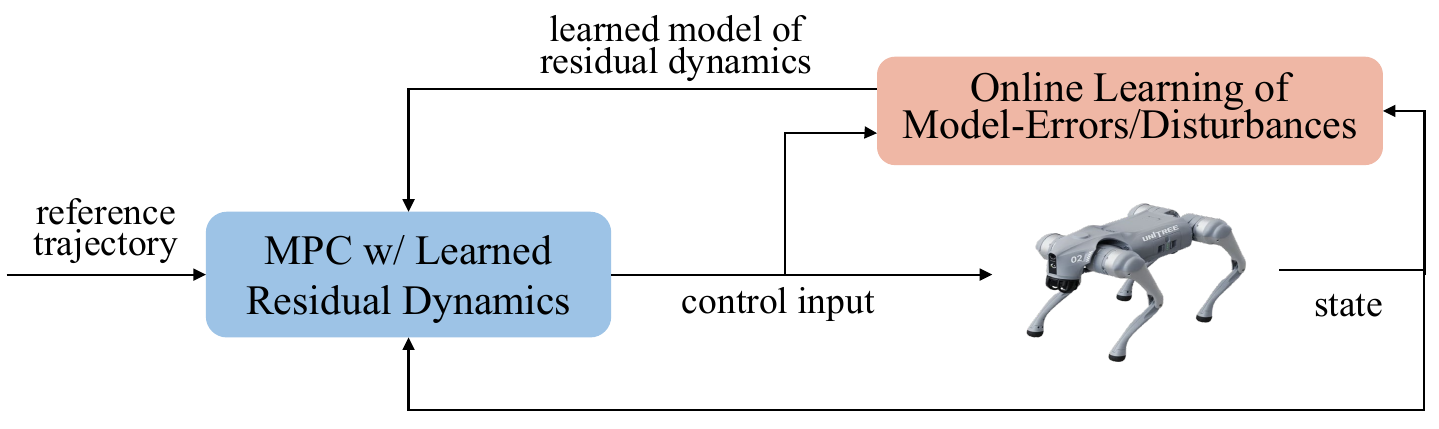}
    \caption{\textbf{Architecture of Adaptive Legged Locomotion via Online Learning and Model Predictive Control.} The pipeline is composed of two modules: (i) a model predictive control (\MPC) module, and (ii) an online learning module. The \MPC module uses the learned residual dynamics model from the online learning module to calculate the next control input. Given the control input and the observed new state, the online learning module then updates the residual dynamics model.}
    \label{fig:framework}
    \vspace{-3mm}
\end{figure}

In this paper, instead, we leverage the success of online learning and model predictive control methods for accurate tracking control under uncertainty~\cite{zhou2024simultaneous}.  
To this end, we learn online a predictive model of the residual dynamics in a self-supervised manner~(Fig.~\ref{fig:framework}). Therefore, the method achieves: one-shot online learning, instead of offline; online control that adapts to the actual disturbance realization, instead of the worst-case; and model predictive control~(\MPC) with a learned residual dynamics model in a reproducing kernel Hilbert space~(\RKHS), instead of vector-valued or linear functions.  We elaborate on our contributions below.

\myParagraph{Contributions} 
We provide a real-time and asymptotically-optimal algorithm for adaptive legged locomotion under unknown residual dynamics.
The algorithm is composed of two interacting modules (Fig.~\ref{fig:framework}): (i) a \MPC module, and (ii) an online learning module.  
At each time step, the \MPC module uses the learned residual dynamics model from the online learning module to calculate the next control input. Given the control input and the observed new state, the online learning module then updates the residual dynamics model.
The update in the online learning module is based on online least-squares estimation via online gradient descent (\OGD)~\cite{hazan2016introduction}, where the residual dynamics in \RKHS~\cite{cucker2002mathematical} are parameterized as a linear combination of random Fourier features~\cite{rahimi2007random,rahimi2008uniform,boffi2022nonparametric}.  
This allows us to learn residual dynamics from a rich functional class while maintaining the computational efficiency of classical finite-dimensional parametric approximations that can be used in \MPC.
The algorithm is asymptotically-optimal in the sense of achieving sublinear \textit{dynamic regret}, defined as the suboptimality against an optimal clairvoyant controller that knows how the unknown residual dynamics.


\myParagraph{Numerical Experiments}
We validate the algorithm in simulated scenarios in Gazebo with the quadruped aiming to track a reference trajectory despite uncertainty up to $12\boldsymbol{g}$, where $\boldsymbol{g}$ is the gravity vector. 
We test the algorithm in flat terrain, slope terrain with $20\degree$ inclination, and rough terrain with $0.25m$ height variation. 
We compare our algorithm with a nominal \MPC that ignores the unknown dynamics or disturbances, and a heuristic L1 \MPC~(\LMPC) that uses an estimated vector value of uncertainty across the \MPC horizon.
The algorithm (i) achieves up to $67\%$ improvement of tracking performance over the nominal \MPC and $21\%$ improvement over \LMPC, and (ii) succeeds even when the nominal \MPC fails, despite the uncertainty due to external forces and challenging terrains.
We validate the algorithm in simulated scenarios in MuJoCo under time-varying uncertainty in flat terrains of different ground coefficients with up to $8~kg$ payload, showing the algorithm achieves significantly better tracking performance than \NMPC.

\vspace{-4mm}
\section{Related Work}\label{sec:lit_review}
We discuss related work on legged control using (i) reinforcement learning and (ii) robust and adaptive control, and related work on online learning for control.

\paragraph*{Reinforcement learning} 
Reinforcement learning algorithms have achieved dynamic legged locomotion by training control policy in a simulator with massive parallelization~\cite{tan2018sim,hwangbo2019learning,lee2020learning,margolis2022walktheseways,gangapurwala2022rloc,kumar2021rma,ji2022concurrent,nahrendra2023dreamwaq,zhong2025bridging}. 
Despite their success, they often face sim-to-real gap when deploying the policy trained in simulator to hardware. 
To address this, a wide range of environment parameters and sensor noises are used to learn the control policy which is robust in this range~\cite{tan2018sim,lee2020learning,margolis2022walktheseways,gangapurwala2022rloc}.
However, domain randomization trades optimality for robustness, leading to a conservative control policy~\cite{luo2017robust}.
Alternatively, a high-fidelity simulator built by using real-world robot data can be used~\cite{tan2018sim,hwangbo2019learning}, but it can be time-consuming and not transferable to different robots.
\cite{kumar2021rma,ji2022concurrent,nahrendra2023dreamwaq,zhong2025bridging} train an environment encoder to obtain environmental information based on onboard sensors in a latent space that is used for the trained policy for adaptation.
In this paper, instead, our algorithm requires no offline data collection and training, achieving one-shot online learning in a self-supervised manner based on data collected online and adapting to real-world environments on-the-spot.

\paragraph*{Robust and adaptive control}
Robust control methods \cite{mayne2011tube,raimondo2009min} select control inputs assuming the worst-case realization of disturbances. However, assuming the worst-case disturbances can be conservative. In addition, these approaches are also computationally expensive, thus limiting their applications to real-time control of legged locomotion~\cite{pandala2022robust,xu2023robust}, where convexification of the \MPC problem or a specialized \MPC solver is required.
Adaptive control methods~\cite{slotine1991applied,krstic1995nonlinear,ioannou1996robust} often assume parametric uncertainty additive to the known system dynamics and update these coefficients online to enhance the robustness against disturbances.
Our method falls into the class of adaptive control methods and learns a model of disturbances online to be used in \MPC. Notably, our method (i) requires no offline system identification of basis functions as required in \cite{minniti2021adaptive}, and (ii) learns a function in a Reproducing Kernel Hilbert Space \emph{(\RKHS)} using random Fourier features, in contrast to a linear function in
\cite{sun2021online,sombolestan2024adaptive,elobaid2025adaptive}.

\paragraph*{Online learning for control} 
Online learning algorithms based on online convex optimization~(\OCO)~\cite{hazan2016introduction,agarwal2019online,zhou2023safecdc,zhou2023efficient,zhou2023saferal,boffi2021regret,tsiamis2024predictive,zhou2024simultaneous,zhou2025no} consider the control problem as a sequential game between a controller and an environment.
They quantify the control performance through \textit{regret}, \ie the suboptimality against an optimal clairvoyant controller that knows the unknown disturbances and dynamics.
\cite{agarwal2019online,zhou2023safecdc,zhou2023efficient,zhou2023saferal} update control inputs based on observed disturbances only since they assume no model that can be used to simulate the future evolution of the disturbances. 
The proposed approaches have been observed to be sensitive to the tuning parameters in \cite{zhou2023safecdc}. 
Instead of optimizing the online controller, \cite{tsiamis2024predictive,zhou2024simultaneous,zhou2025no} learn the model of the disturbances, and use model predictive control to select control inputs based on the learned disturbance model.

\section{Adaptive Legged Locomotion via Online Learning and Model Predictive Control}\label{sec:problem}

We formulate the problem of \textit{Adaptive Legged Locomotion via Online Learning and Model Predictive Control}~(\Cref{prob:control}).
We use the following framework and assumptions.

\myParagraph{Single-Rigid Body Dynamics}
We consider the single-rigid body dynamics of the form

\begin{equation}
    \begin{aligned}
        \dot{\boldsymbol{p}} &= \boldsymbol{v}, \; & m \dot{\boldsymbol{v}} &= m \boldsymbol{g} + \boldsymbol{R}\sum_{i=1}^{4}\boldsymbol{f}_{i} + \boldsymbol{f}_{u},  \\
        \dot{\boldsymbol{\theta}} &= \boldsymbol{T}\left(\boldsymbol{\theta}\right) \boldsymbol{\omega}, \; & \boldsymbol{\mathcal{J}} \dot{\boldsymbol{\omega}} &= -\boldsymbol{\omega} \times \boldsymbol{\mathcal{J}} \boldsymbol{\omega} + \sum_{i=1}^{4}\boldsymbol{r}_{i}  \times \boldsymbol{f}_{i} + \boldsymbol{\tau}_{u}, 
    \end{aligned}
    \label{eq:srb_dynamics}
\end{equation}
where $\boldsymbol{p} \in \mathbb{R}^{3}$ and $\boldsymbol{v} \in \mathbb{R}^{3}$ are position and velocity in the inertial frame, $\boldsymbol{\theta}$ is the Euler angle, $\boldsymbol{\omega} \in \mathbb{R}^{3}$ is the body angular velocity, $m$ is the mass, $\boldsymbol{\mathcal{J}}$ is the inertia matrix, $\boldsymbol{g}$ is the gravity vector, $\boldsymbol{R}\in SO(3)$ is the rotation matrix from the body to inertial frame, $\boldsymbol{T}: \mathbb{R}^{3} \rightarrow \mathbb{R}^{3 \times 3}$ is the Euler angle transformation matrix, $\boldsymbol{f}_i \in \mathbb{R}^3$ is the contact force of the $i$-th foot, $\boldsymbol{r}_i \in \mathbb{R}^3$ is the $i$-th foot's position in body frame, $\boldsymbol{f}_{u} \in \mathbb{R}^3$ is the unknown force in inertial frame, and $\boldsymbol{\tau}_{u} \in \mathbb{R}^3$ is the unknown torque in body frame. 

For convenience, we rewrite \cref{eq:srb_dynamics} into the form of a control-affine system as follows:
\begin{equation}
    \dot{\boldsymbol{x}} = \boldsymbol{f}\left(\boldsymbol{x}\right) + 
    \left[\begin{array}{c}
         \boldsymbol{0}_{6\times1}  \\
         {\boldsymbol{J}^{\top}}
    \end{array}\right] \boldsymbol{u} + \left[\begin{array}{c}
         \boldsymbol{0}_{6\times1}  \\
         \boldsymbol{h} \left(\boldsymbol{z}\right)
    \end{array}\right],
    \label{eq:affine_sys_cont}
\end{equation}
where $\boldsymbol{x} \triangleq\left[\boldsymbol{p}^\top \; \boldsymbol{\theta}^\top \; \boldsymbol{v}^\top \; \boldsymbol{\omega}^\top \right]^\top \in\mathbb{R}^{12}$ is the state, $\boldsymbol{u} \triangleq\left[\boldsymbol{f}_{1}^\top \; \boldsymbol{f}_{2}^\top \; \boldsymbol{f}_{3}^\top \; \boldsymbol{f}_{4}^\top \right]^\top\in\mathbb{R}^{12}$ is the control input, $\boldsymbol{f}: \mathbb{R}^{12} {\rightarrow} \mathbb{R}^{12}$ is a  known locally Lipschitz function, {$\boldsymbol{J} \in \mathbb{R}^{12\times 6}$} is the contact Jacobian matrix that depends on $\boldsymbol{x}$ and $\boldsymbol{r}_{i}$, $\boldsymbol{h}\triangleq\left[\boldsymbol{f}_{u}^\top \; \boldsymbol{\tau}_{u}^\top \right]^\top: \mathbb{R}^{d_z} \rightarrow \mathbb{R}^{6}$ is the unknown disturbances, and $\boldsymbol{z} \in\mathbb{R}^{d_z}$ is a vector of features chosen as a subset of $[\boldsymbol{x}^\top \ \boldsymbol{u}^\top]^\top$.  $\boldsymbol{h}\left(\cdot\right)$ represents unknown residual dynamics that depend on system state and control input, which can be used to model the effects of unknown payload and uneven terrains.

Using forward Euler discretization, we can obtain the discrete-time system dynamics from \cref{eq:affine_sys_cont}:
\begin{equation}
    \boldsymbol{x}_{t+1} = \boldsymbol{f}\left(\boldsymbol{x}_{t}\right) + 
    \left[\begin{array}{c}
         \boldsymbol{0}_{6\times1}  \\
         {\boldsymbol{J}_{t}^{\top}}
    \end{array}\right] \boldsymbol{u}_{t} + \left[\begin{array}{c}
         \boldsymbol{0}_{6\times1}  \\
         \boldsymbol{h} \left(\boldsymbol{z}_{t}\right)
    \end{array}\right],
    \label{eq:affine_sys_disc}
\end{equation}
where we overload the notations used in \cref{eq:affine_sys_cont} and omit the discretization time.
We refer to the undisturbed system dynamics as the \textit{nominal dynamics}.

\myParagraph{Model Predictive Control (\MPC)} 
\MPC selects a control input $\boldsymbol{u}_t$ by simulating the system dynamics over a look-ahead horizon $N$. In the presence of unknown residual dynamics, \MPC can utilize an estimate of $\boldsymbol{h}\left(\cdot\right)$:
\begin{subequations}
    \label{eq:mpc_ada_def}
    \begin{align}
        &\underset{{\boldsymbol{x}_{t+1:t+N}, \; \boldsymbol{u}_{t:t+N-1}}}{\textit{min}} \sum_{k=t}^{t+N-1} c_{k}\left(\boldsymbol{x}_{k},\boldsymbol{u}_{k}\right) \label{eq:mpc_ada_def_obj} \\
        & \ \ \operatorname{\textit{subject~to}} \;\quad \boldsymbol{x}_{k+1} = \boldsymbol{f}\left(\boldsymbol{x}_{k}\right) + 
        \left[\begin{array}{c}
             \boldsymbol{0}_{6\times1}  \\
             {\boldsymbol{J}_{k}^{\top}}
        \end{array}\right] \boldsymbol{u}_{k} + \left[\begin{array}{c}
         \boldsymbol{0}_{6\times1}  \\
         \hat{\boldsymbol{h}} \left(\boldsymbol{z}_{k}\right)
        \end{array}\right],\label{eq:mpc_ada_dyn}\\
        & \qquad \qquad \qquad \; \boldsymbol{u}_{k}\in \calU, \ \ k\in\{t,\ldots, t+N-1\},
    \end{align}
\end{subequations}
where $c_{t}\left(\cdot,\cdot\right): \mathbb{R}^{12} \times \mathbb{R}^{12} {\rightarrow} \mathbb{R}$ is the cost function, $\calU$ is a compact set that represents constraints on the control input due to, \eg controller saturation and friction cone,
$\hat{\boldsymbol{h}}\left(\cdot\right)$ is the estimate of $\boldsymbol{h}\left(\cdot\right)$.  Specifically, $\hat{\boldsymbol{h}}\left(\cdot\right)  \triangleq \hat{\boldsymbol{h}}\left(\cdot~; \hat{\boldsymbol{\alpha}}\right)$ where $\hat{\boldsymbol{\alpha}}$ is a parameter that is updated online by our proposed method to improve the control performance.

\myParagraph{Control Performance Metric} We design $\boldsymbol{u}_t$ to ensure a control performance that is comparable to an optimal clairvoyant (non-causal) policy that knows the disturbance function $\boldsymbol{h}$ a priori. Particularly, we consider the metric below.

\begin{definition}[Dynamic Regret]\label{def:DyReg_control}
Assume a total time horizon of operation $T$, and loss functions $c_t$, $t=1,\ldots, T$. Then, \emph{dynamic regret} is defined as
\begin{equation}
	\DReg = \sum_{t=1}^{T} c_{t}\left(\boldsymbol{x}_{t}, \boldsymbol{u}_{t}, \boldsymbol{h}(\boldsymbol{z}_t)\right)-\sum_{t=1}^{T} c_{t}\left(\boldsymbol{x}_{t}^{\star}, \boldsymbol{u}_{t}^{\star}, \boldsymbol{h}(\boldsymbol{z}_t^\star)\right),
	\label{eq:DyReg_control}
\end{equation}
where we made the dependence of the cost $c_t$ to the unknown disturbance $\boldsymbol{h}$ explicit, $\boldsymbol{u}_{t}^{\star}$ is the optimal control input in hindsight, \ie the optimal (non-causal) input given a priori knowledge of the unknown function $\boldsymbol{h}$ and $\boldsymbol{x}_{t+1}^{\star}$ is the state reached by applying the optimal control inputs $\left(\boldsymbol{u}_{1}^{\star}, \; \dots, \; \boldsymbol{u}_{t}^{\star}\right)$.
\end{definition}


\begin{problem}[Adaptive Legged Locomotion via Online Learning and Model Predictive Control]\label{prob:control}
At each $t=1,\ldots, T$, estimate the unknown dynamics $\hat{\boldsymbol{h}}\left(\cdot\right)$, and identify a control input $\boldsymbol{u}_t$ by solving \cref{eq:mpc_ada_def}, such that $\DReg$ is sublinear.
\end{problem}

A sublinear dynamics regret means $\lim_{T\rightarrow\infty} \DReg/T \rightarrow 0$, which implies the algorithm asymptotically converges to the optimal (non-causal) controller.


\section{Algorithm and Regret Guarantee}\label{sec:alg}

We present the algorithm for \Cref{prob:control} (\Cref{alg:MPC}) and its performance guarantee. The algorithm is sketched in \Cref{fig:framework}.  The algorithm is composed of two interacting modules: (i) an \MPC module, and (ii) an online system identification module.  At each $t=1,2,\ldots,$ the \MPC module uses the estimated $\hat{\boldsymbol{h}}(\cdot)$ from the system identification module to calculate the control input $\boldsymbol{u}_t$. Given the current control input $\boldsymbol{u}_t$ and the observed new state $\boldsymbol{x}_{t+1}$, the online system identification module updates the estimate $\hat{\boldsymbol{h}}(\cdot)$.  To this end, it employs online least-squares estimation via online gradient descent, where $\boldsymbol{h}(\cdot)$ is parameterized as a linear combination of random fourier features. 
To rigorously present the algorithm, we thus first introduce random Fourier features for approximating an $\boldsymbol{h}\left(\cdot\right)$~(\Cref{subsec:RFF}), and online gradient descent for estimation~(\Cref{subsec:OLS}).

\subsection{Function Approximation via Random Fourier Features}\label{subsec:RFF}

We overview the randomized approximation algorithm in~\cite{boffi2022nonparametric} for approximating an $\boldsymbol{h}\left(\cdot\right)$.
The algorithm is based on random Fourier features~\cite{rahimi2007random,rahimi2008uniform} and their extension to vector-valued functions~\cite{brault2016random,minh2016operator}.
By being randomized, the algorithm is computationally efficient
while retaining the expressiveness of the \RKHS with high probability.

Based on the assumptions of $\boldsymbol{h}: \mathbb{R}^{d_z} \rightarrow \mathbb{R}^{d_x}$ lies in a subspace of a Reproducing Kernel Hilbert Space \emph{(\RKHS)} $\calH$~\cite{bach2017breaking} and the Operator-Valued Bochner's Theorem~\cite{brault2016random}, 
we assume that $\boldsymbol{h}$ can be written as $\boldsymbol{h}\left(\cdot\right) = \int_\Theta \boldsymbol{\Phi}\left(\cdot, \boldsymbol{\theta}\right) \boldsymbol{\alpha}(\boldsymbol{\theta}) \mathrm{d}\nu (\boldsymbol{\theta})$,~\cite{bach2017breaking}
and there exists a finite-dimensional approximation of $\boldsymbol{h}\left(\cdot\right)$ by $\boldsymbol{h}\left(\cdot\right) \approx \hat{\boldsymbol{h}}(\cdot;\boldsymbol{\alpha})\triangleq\frac{1}{M} \sum_{i=1}^{M} \boldsymbol{\Phi}\left(\cdot, \boldsymbol{\theta}_i \right) \boldsymbol{\alpha}_i,$
where  $\boldsymbol{\Phi} \left(\boldsymbol{z}, \boldsymbol{\theta}\right)= \boldsymbol{B}(\boldsymbol{w}) \phi\left(\boldsymbol{w}^{\top} \boldsymbol{z}+b\right)$ is the feature map, 
$\boldsymbol{B}: \mathbb{R}^{d_z} \rightarrow \mathbb{R}^{d_x \times d_{1}}$, $\phi: \mathbb{R} \rightarrow[-1,1]$ is a $1$-Lipschitz function, 
$d_1 \leq d_x$, 
$\boldsymbol{\theta}_i \sim \nu$ are drawn i.i.d. from the base measure $\nu$ with $\boldsymbol{\theta}=\left(\boldsymbol{w}, b\right)$, $\boldsymbol{w} \in \mathbb{R}^{d_z}$, and $b \in \mathbb{R}$, 
$\boldsymbol{\alpha}_i \triangleq \boldsymbol{\alpha}\left(\theta_i\right)$ are parameters to be learned, and $M$ is the number of sampling points that decides the approximation accuracy.

The following shows the expressiveness of the  finite-dimensional approximation of $\boldsymbol{h}\left(\cdot\right)$, considering $\boldsymbol{\alpha}_{i} \in \calD$, where 
$\calD\triangleq \{ \boldsymbol{\alpha} \mid \|\boldsymbol{\alpha}\| \leq B_h\}$.

\begin{proposition}[Uniformly Approximation Error~\cite{boffi2022nonparametric}]\label{prop:approx_error}
     Assume $\boldsymbol{h} \in$ $\mathcal{F}_{2}\left(B_{{h}}\right)$, where $$\calF_2 \left(B_h\right) \triangleq  \Bigg\{ \boldsymbol{h}\left(\cdot\right) = \left. \int_\Theta \boldsymbol{\Phi}\left(\cdot, \boldsymbol{\theta}\right) \boldsymbol{\alpha}(\boldsymbol{\theta}) \mathrm{d}\nu (\boldsymbol{\theta})  \right\vert \boldsymbol{\alpha} \in \calD \Bigg\}.$$ 
     Let $\delta \in(0,1)$.  With probability at least $1-\delta$, there exist $\left\{\boldsymbol{\alpha}_{i}\right\}_{i=1}^{M} \in \calD$, \ie $\|\boldsymbol{\alpha}_{i}\| \leq B_h$, such that
\begin{equation} 
        \left\| \boldsymbol{h}\left(\cdot\right) - \frac{1}{M} \sum_{i=1}^{M} \boldsymbol{\Phi}\left(\cdot, \boldsymbol{\theta}_{i}\right) \boldsymbol{\alpha}_{i}\right\|_{\infty} \leq \calO\left(\frac{1}{\sqrt{M}}\right).
    \end{equation}
\end{proposition}
    
\Cref{prop:approx_error}, therefore, indicates that the uniformly approximation error scales $\calO\left(\frac{1}{\sqrt{M}}\right)$.

Random Fourier features can be viewed as linearizations of neural networks~\cite{ghorbani2021linearized,jacot2018neural}. Neural networks, in principle, can perform better than kernel methods due to greater expressivity.  However, using neural networks poses challenges in such online learning settings due to their data-hungry nature. In addition, using neural networks in \MPC can be computationally expensive for embedded systems, and a customized solver or dynamics representation is required~\cite{salzmann2023real,saviolo2023active}.
Therefore, we utilize random Fourier features to balance computational efficiency and expressiveness.

\subsection{Online Least-Squares Estimation}\label{subsec:OLS}
Given a data point $\left( \boldsymbol{z}_{t}, \; \boldsymbol{h}\left(\boldsymbol{z}_{t}\right) \right)$ observed at time $t$, we employ an online least-squares algorithm that updates the parameters $\hat{\boldsymbol{\alpha}}_t \triangleq \left[ \boldsymbol{\alpha}_{i,t}^\top, \; \dots, \; \boldsymbol{\alpha}_{M,t}^\top\right]^\top$ to minimize the approximation error $l_t = \| \boldsymbol{h}\left(\boldsymbol{z}_{t}\right) - \hat{\boldsymbol{h}}\left(\boldsymbol{z}_{t}\right) \|^2$, where $ \hat{\boldsymbol{h}}(\cdot) \triangleq  \frac{1}{M} \sum_{i=1}^{M} \boldsymbol{\Phi}\left(\cdot, \boldsymbol{\theta}_i \right) \hat{\boldsymbol{\alpha}}_{i,t} $ and $\boldsymbol{\Phi}\left(\cdot,\boldsymbol{\theta}_i\right) $ is the random Fourier feature as in \Cref{subsec:RFF}. 
Specifically, the algorithm used the online gradient descent algorithm~(\OGD)~\cite{hazan2016introduction}. At each $t = 1, \dots, T$, it makes the steps:
\begin{itemize}
    \item Given $\left( \boldsymbol{z}_{t}, \; \boldsymbol{h}\left(\boldsymbol{z}_{t}\right) \right)$, formulate the estimation loss function (approximation error):
            \begin{equation*}
                l_t\left(\hat{\boldsymbol{\alpha}}_t\right) \triangleq \left\| \boldsymbol{h}\left(\boldsymbol{z}_{t}\right) -  \frac{1}{M} \sum_{i=1}^{M} \boldsymbol{\Phi}\left(\boldsymbol{z}_{t}, \boldsymbol{\theta}_i \right) \hat{\boldsymbol{\alpha}}_{i,t} \right\|^2.
            \end{equation*}
    \item Calculate the gradient of $l_t\left(\hat{\boldsymbol{\alpha}}_t\right)$ with respect to $\hat{\boldsymbol{\alpha}}_t$: 
            \begin{equation*}
                \nabla_t \triangleq \nabla_{\hat{\boldsymbol{\alpha}}_t} l_t\left(\hat{\boldsymbol{\alpha}}_t\right).
            \end{equation*}
    \item Update using gradient descent with learning rate $\eta$:
            \begin{equation*}
                \hat{\boldsymbol{\alpha}}_{t+1}^\prime= \hat{\boldsymbol{\alpha}}_t- \eta \nabla_t.
            \end{equation*}
    \item Project each $\hat{\boldsymbol{\alpha}}_{i,t+1}^\prime$ onto $\calD$:
            \begin{equation*}
                \hat{\boldsymbol{\alpha}}_{i,t+1} = \Pi_{\calD}(\hat{\boldsymbol{\alpha}}_{i,t+1}^\prime) \triangleq \underset{\boldsymbol{\alpha} \in \calD}{\operatorname{\textit{argmin}}}\; \| \boldsymbol{\alpha} - \hat{\boldsymbol{\alpha}}_{i,t+1}^\prime \|^2.
            \end{equation*}
\end{itemize}

The above online least-squares estimation algorithm enjoys an $\calO\left(\sqrt{T}\right)$ regret bound, per the regret bound of \OGD~\cite{hazan2016introduction}.

\begin{proposition}[Regret Bound of Online Least-Squares Estimation~\cite{hazan2016introduction}]\label{theorem:OGD}
    Assume $\eta=\calO\left({1}/{\sqrt{T}}\right)$.  Then,
    \begin{equation}
       \SReg\triangleq \sum_{t=1}^{T} l_t \left(\boldsymbol{\alpha}_t\right) - \sum_{t=1}^{T} l_t \left(\boldsymbol{\alpha}^{\star}\right)  \leq \calO\left(\sqrt{T}\right),
    \end{equation}
    where $\boldsymbol{\alpha}^{\star} \triangleq {\operatorname{\textit{argmin}}}\;\sum_{t=1}^{T} l_t \left(\boldsymbol{\alpha}\right)$ is the optimal parameter that achieves lowest cumulative loss in hindsight.
\end{proposition}

The online least-squares estimation algorithm thus asymptotically achieves the same estimation error 
as the optimal parameter $\boldsymbol{\alpha}^{\star}$ since $\lim_{T\rightarrow\infty} \;\SReg/T = 0$. 


\subsection{Algorithm for \Cref{prob:control}}\label{subsec:MPC}

\begin{algorithm}[t]
	\caption{Adaptive Legged Locomotion via Online Learning and Model Predictive Control.}
	\begin{algorithmic}[1]
		\REQUIRE Number of random Fourier features $M$; base measure $\nu$; domain set $\calD$;  gradient descent learning rate $\eta$.
		\ENSURE Ground reaction forces $\boldsymbol{u}_{t}$.
		\medskip
            \STATE Initialize $\boldsymbol{x}_1$, $\hat{\boldsymbol{\alpha}}_{i,1} \in \calD$; 
            \STATE Randomly sample $\boldsymbol{\theta}_i \sim \nu$ and formulate $\boldsymbol{\Phi}\left(\cdot, \boldsymbol{\theta}_i\right)$, where $i \in \{1, \dots, M\}$;
		\FOR {each time step $t = 1, \dots, T$}
		\STATE Receive contact schedule, desired foothold positions, and reference trajectory;
		\STATE Formulate \cref{eq:mpc_ada_def} with $\hat{\boldsymbol{h}}(\cdot) \triangleq  \frac{1}{M} \sum_{i=1}^{M} \boldsymbol{\Phi}\left(\cdot, \boldsymbol{\theta}_i \right) \hat{\boldsymbol{\alpha}}_{i,t} $;
		\STATE Obtain ground reaction forces $\boldsymbol{u}_t$ by solving \cref{eq:mpc_ada_def} and send $\boldsymbol{u}_t$ to low-level leg controller;
            \STATE Observe state $\boldsymbol{x}_{t+1}$, and calculate $\boldsymbol{h}\left(\boldsymbol{z}_t\right)$ via \cref{eq:affine_sys_disc};
            \STATE Formulate estimation loss $l_t\left(\hat{\boldsymbol{\alpha}}_t\right) \triangleq \| \boldsymbol{h}\left(\boldsymbol{z}_t\right) - \frac{1}{M} \sum_{i=1}^{M} \boldsymbol{\Phi}\left(\boldsymbol{z}_t, \boldsymbol{\theta}_i \right) \hat{\boldsymbol{\alpha}}_{i,t} \|^2$;
            \STATE Calculate gradient $\nabla_t \triangleq \nabla_{\hat{\boldsymbol{\alpha}}_t} l_t\left(\hat{\boldsymbol{\alpha}}_t\right)$;
            \STATE Update $\hat{\boldsymbol{\alpha}}_{t+1}^\prime= \hat{\boldsymbol{\alpha}}_t- \eta \nabla_t$;
            \STATE Project  $\hat{\boldsymbol{\alpha}}_{i,t+1}^\prime$ onto $\calD$, \ie $\hat{\boldsymbol{\alpha}}_{i,t+1} = \Pi_{\calD}(\hat{\boldsymbol{\alpha}}_{i,t+1}^\prime)$, for $i \in \{1, \; \dots, \; M\}$;
            \ENDFOR
	\end{algorithmic}\label{alg:MPC}
\end{algorithm}

The pseudo-code is given in \Cref{alg:MPC}.  
The algorithm is composed of three steps, initialization, control, and online learning, where the control and online learning steps influence each other at each time steps~(\Cref{fig:framework}):

\begin{itemize}
    \item \textit{Initialization steps:} \Cref{alg:MPC} first initializes the system state $\boldsymbol{x}_1$ and parameter $\hat{\boldsymbol{\alpha}}_1 \in \calD$~(line 1). Then given the number of random Fourier features, \Cref{alg:MPC} randomly samples $\theta_i$ and formulates $\boldsymbol{\Phi}\left(\cdot, \boldsymbol{\theta}_i\right)$, where $i \in \{1, \dots, M\}$~(line 2).

    \item \textit{Control steps:} Then, at each~$t$, given the current estimate $\hat{\boldsymbol{h}}(\cdot) \triangleq  \frac{1}{M} \sum_{i=1}^{M} \boldsymbol{\Phi}\left(\cdot, \boldsymbol{\theta}_i \right) \hat{\boldsymbol{\alpha}}_{i,t}$, contact schedule, desired foothold positions, and reference trajectory, \Cref{alg:MPC} applies the ground reaction forces $\boldsymbol{u}_t$ obtained by solving \cref{eq:mpc_ada_def}~(lines 4-6).

    \item \textit{Learning steps:}
    The system then evolves to state $\boldsymbol{x}_{t+1}$, and, $\boldsymbol{h}\left(\boldsymbol{z}_{t}\right)$ is calculated upon observing $\boldsymbol{x}_{t+1}$~(line 7).
    Afterwards, the algorithm formulates the loss $l_t\left(\hat{\boldsymbol{\alpha}}_t\right) \triangleq \| \boldsymbol{h}\left(\boldsymbol{z}_{t}\right) - \sum_{i=1}^{M} \boldsymbol{\Phi}\left(\boldsymbol{z}_{t}, \boldsymbol{\theta}_i \right) \hat{\boldsymbol{\alpha}}_{i,t} \|^2$, and calculates the gradient $\nabla_t \triangleq \nabla_{\hat{\boldsymbol{\alpha}}_t} l_t\left(\hat{\boldsymbol{\alpha}}_t\right)$~(lines 8-9). 
    \Cref{alg:MPC} then updates the parameter $\hat{\boldsymbol{\alpha}}_t$ to $\hat{\boldsymbol{\alpha}}_{t+1}^\prime$~(line 10) and, finally, projects each $\hat{\boldsymbol{\alpha}}_{i,t+1}^\prime$ back to the domain set $\calD$~(line 11).
\end{itemize}

\subsection{No-Regret Guarantee}\label{sec:Reg}

We present the sublinear regret bound of \Cref{alg:MPC}~\cite[Theorem~1]{zhou2024simultaneous}. 

\begin{theorem}[Dynamic Regret Guarantee~\cite{zhou2024simultaneous}]\label{theorem:regret_OLMPC}
Assume $\eta=\calO\left({1}/{\sqrt{T}}\right)$ in \Cref{alg:MPC}. Consider the dynamics in \cref{eq:affine_sys_disc} corrupted with unmodeled noise $\boldsymbol{e}_t$.
\Cref{alg:MPC} achieves 
\begin{equation}
    \DReg \leq \calO\left(T^{\frac{3}{4}}\right) + \calO\left( \sqrt{ T\sum_{t=1}^{T}\|\boldsymbol{e}_{t}\|^2 }\right).
    \label{eq:theorem_regret_OLMPC}
\end{equation}
\end{theorem}

In the absence of $\boldsymbol{e}_t$, which we consider in the paper, \Cref{theorem:regret_OLMPC} reduces to $\DReg \leq \calO\left(T^{\frac{3}{4}}\right)$. Therefore, \Cref{theorem:regret_OLMPC} serves as a finite-time performance guarantee as well as implies that \Cref{alg:MPC} converges to the optimal (non-causal) control policy since $\lim_{T\rightarrow\infty}\DReg / T \rightarrow 0$. 

The bound holds under the assumptions of stability of the estimated systems, Lipschitzness of ${c}_t\left(\boldsymbol{x},\boldsymbol{u}\right)$ in $\boldsymbol{x}$ and $\boldsymbol{u}$, Lipschitzness of $\hat{\boldsymbol{h}}\left(\cdot\right)$ in $\hat{\boldsymbol{\alpha}}$, and $\boldsymbol{h}\left(\cdot\right)$ can be expressed as $\frac{1}{M} \sum_{i=1}^{M} \boldsymbol{\Phi}\left(\cdot, \boldsymbol{\theta}_{i}\right) \boldsymbol{\alpha}_{i}$. We refer readers to \cite{zhou2024simultaneous} for detailed statements of the assumptions.

\section{Numerical Experiments}\label{sec:sim}
We evaluate \Cref{alg:MPC} in simulated scenarios of legged control under uncertainty, where the quadruped aims to track a reference trajectory despite unknown external disturbances. 
We conduct experiments on Gazabo~(\Cref{subsec:sim-gazebo}) and MuJoCo~(\Cref{subsec:sim-mujoco}) simulators.
We detail the simulation setup and results below.

\begin{table*}[t]
    \centering
    \caption{\textbf{Performance Comparison for the Gazebo Simulations in \Cref{subsec:sim-gazebo}.} The table reports the average value of tracking error in position ($cm$). The \blue{blue} numbers correspond to the \blue{better overall performance}. Failure is denoted by $-$. Our method achieves better tracking performance than nominal \scenariot{MPC}.}
    \label{table:sim}
    \resizebox{2\columnwidth}{!}{
    {
    \begin{tabular}{cccccccccccccc}
    \toprule
    \multirow{2}{*}{Terrain} & \multirow{2}{*}{$\boldsymbol{f}_{u}$ ($N$)} & \multicolumn{3}{c}{$x$} & \multicolumn{3}{c}{$y$} & \multicolumn{3}{c}{$z$} & \multicolumn{3}{c}{Overall}  \cr
    \cmidrule(lr){3-5} \cmidrule(lr){6-8} \cmidrule(lr){9-11} \cmidrule(lr){12-14} & & Nominal & L1 & Ours & Nominal & L1 & Ours & Nominal & L1 & Ours & Nominal & L1 & Ours \cr
    \midrule
    
    \multirow{4}{*}{Flat}  & \multirow{4}{*}{$\begin{array}{c} \boldsymbol{0} \\ 4\boldsymbol{g} \\  8\boldsymbol{g} \\ 12\boldsymbol{g} \end{array}$} &
    $ 2.23 $ & $ 2.64$ &  $ 2.70 $  &   $ 0.31 $  & $ 0.36$ &    $ 0.32 $ & $ 0.84 $   & $ 0.51$ & $ 0.51$ & \blue{$ 2.51 $}  & $ 2.79$ & $ 2.85 $ \cr
    & & $ 2.02 $ & $2.70 $ &  $ 2.89 $  &   $ 0.41 $ & $0.40 $  &    $ 0.41 $  & $ 5.59  $ & $1.04 $ &  $ 1.02 $  &  $ 6.07$  & \blue{$ 3.33 $} &  $ 3.39 $  \cr
    & & $ 1.04 $  & $ 2.88$ &  $ 3.06$  &   $ 1.66$  & $ 0.57$ &    $ 0.60 $  & $ 11.60$  & $ 2.68$ &  $ 2.10 $  &  $ 11.87$  & $ 4.61 $ &  \blue{$ 4.30 $}  \cr
    & & -  & $ 2.99 $ &  $ 3.58$  &   -  & $ 0.84 $ &    $ 0.87$  & -  & $ 4.58$ &  $ 3.00$  &  -  & $ 6.17 $ &  \blue{$ 5.69$}  \cr
    \midrule
    
    \multirow{4}{*}{Slope}  & \multirow{4}{*}{$\begin{array}{c} \boldsymbol{0} \\ 4\boldsymbol{g} \\  8\boldsymbol{g} \\ 12\boldsymbol{g} \end{array}$} &
    $ 2.86 $ & $ 2.78$ &  $ 2.33$  &   $ 0.57$  & $ 0.52$ &    $0.54$ & $ 1.83$    & $ 1.85$ & $ 1.82$ & $ 3.58$  & $3.50$ & \blue{$ 3.12$} \cr
    & & $ 3.01 $  & $ 2.73 $ &  $ 2.46$  &   $ 0.89$  & $ 0.72$ &    $ 0.75$  & $ 7.49$  & $ 2.85$ &  $ 2.37 $  &  $ 8.24$  & $ 4.29$ &  \blue{$ 3.66$}  \cr
    & & $ 1.91$ & $ 2.65$ &  $ 2.62$  &   $ 1.39$  & $ 1.09$ &    $ 1.05$  & $  13.11$  & $ 4.09$ &  $ 3.00$  &  $ 13.43$  & $ 5.40 $ &  \blue{$ 4.44$}  \cr
    & & $  - $  & $ 3.21$ &  $ 3.85$  &   $  - $ & $ 2.87$  &    $ 2.99$  & $  - $ & $ 6.96$   &  $ 6.10$  &  $  - $ & $ 8.73$ &  \blue{$ 8.44 $}  \cr
    \midrule
    
    \multirow{3}{*}{Rough}  & \multirow{3}{*}{$\begin{array}{c} \boldsymbol{0}  \\ 2\left[\|\boldsymbol{g}\|,\;0,\;\|\boldsymbol{g}\|\right]^\top\\ 4\boldsymbol{g}  \end{array}$} &
    $ 0.93$  & $ 1.00$ &  $ 0.95$  &   $ 2.32$  & $ 1.69$ &    $  1.73$  & $ 1.30$  & $ 1.00$ &  $ 1.04$  &  $ 3.14$  & \blue{$ 2.46 $} &  $ 2.52$  \cr
    & & $  - $ & $ 0.94$ &  $ 0.98$  &   $  - $ & $ 3.38$ &    $ 2.47$  & $  - $ & $ 1.35$ &  $ 1.24$  &  $  - $ & $ 4.13$ &  \blue{$ 3.26$}  \cr
    & & $ 0.91$  & $ 0.95$ &  $ 0.96$  &   $ 2.20$  &  $ 2.97$ &  $ 2.47 $  & $ 5.35$  & $ 1.67$ &  $ 1.62$  &  $ 6.10$  &  $ 3.93$ &  \blue{$ 3.45$}  \cr
    \bottomrule
    \end{tabular}}
    }
\end{table*}

\subsection{Gazebo Simulations}\label{subsec:sim-gazebo}

\myParagraph{Simulation Setup} 
We employ the Unitree Go2 Robot in the Quad-SDK Gazebo environment~\cite{abs:norby-quad-sdk-2022}.
The \MPC runs at $200Hz$.
It takes the contact scheduling, desired foothold positions, and reference trajectory as input, and outputs the desired ground reaction forces to the low-level leg controller.
The low-level leg controller runs at $500Hz$.

\myParagraph{Control Design} 
The \MPC uses look-ahead horizon $N=20$ simulating the dynamics for $0.6s$. We use quadratic cost functions with 
$\boldsymbol{Q} = \diag{\boldsymbol{Q}_{\boldsymbol{p}},\; \boldsymbol{Q}_{\boldsymbol{\theta}},\; \boldsymbol{Q}_{\boldsymbol{v}},\; \boldsymbol{Q}_{\boldsymbol{\omega}}}$, 
$\boldsymbol{Q}_{\boldsymbol{p}}=12.5\textbf{I}_{3}$,
$\boldsymbol{Q}_{\boldsymbol{\theta}}=\diag{[0.5,\; 0.5,\; 2.5]}$,
$\boldsymbol{Q}_{\boldsymbol{v}}=\diag{[0.2,\; 0.2,\; 0.4]}$,
$\boldsymbol{Q}_{\boldsymbol{\omega}}=\diag{[0.1,\; 0.1,\; 0.4]}$,
and $\boldsymbol{R}=5e^{-5}\textbf{I}_{12}$.
We use the forward Euler method for discretization. 
We use as the feature $\boldsymbol{z}_t$ the $\boldsymbol{v}_t$, $\boldsymbol{\theta}_t$, $\boldsymbol{\omega}_t$, and ${\boldsymbol{J}_{t}^{\top}} \boldsymbol{u}_t$.
We sample $\boldsymbol{w}_i$ from a Gaussian distribution with standard deviation $0.01$.
We use $M=50$ random Fourier features and $\eta=0.003$, and initialize $\hat{\boldsymbol{\alpha}}$ as a zero vector. 
We do not specify $B_h$ for the domain set $\calD$, and the projection step is not applied.
The nonlinear program in \cref{eq:mpc_ada_def} is constructed by CasADi~\cite{andersson2019casadi} and solved by IPOPT~\cite{wachter2006implementation}.

\myParagraph{Benchmark Experiment Setup}
We consider three types of terrains: flat terrain, slope terrain with $20\degree$ inclination, and rough terrain with $0.25m$ height variation, shown in \Cref{fig:sim-exp}~(third \& fourth).
The quadruped is tasked to walk from position $\left[0,\;0\right]$ to $\left[6,\;0\right]$ while maintaining height of $0.3m$ above the ground. 
In flat and slope terrains, the quadruped walks at $0.75m/s$ with $\boldsymbol{f}_{u} = \boldsymbol{0}$, $4\boldsymbol{g}$, $8\boldsymbol{g}$, and $12\boldsymbol{g}$.
In rough terrain, the quadruped walks at $0.5m/s$ with $\boldsymbol{f}_{u} = \boldsymbol{0}$, $ 2\left[\|\boldsymbol{g}\|,\;0\;,\|\boldsymbol{g}\|\right]^\top$, and $4\boldsymbol{g}$.
We use the tracking error in position as the performance metric.

\myParagraph{Results} 
We compare \Cref{alg:MPC} with a nominal \MPC that assumes no uncertainty in the model, 
and a heuristic L1 \MPC~(\LMPC) based on~\cite{wu2025l1quad}.
Specifically, at each time step, \LMPC first uses L1 adaptation law to estimate a vector value of $\bar{\boldsymbol{h}}$~\cite[Algorithm~1]{wu2025l1quad}.
Then, \LMPC solves \cref{eq:mpc_ada_def} by setting $\hat{\boldsymbol{h}}(\boldsymbol{z}_{k}) = \bar{\boldsymbol{h}}$ for  $k\in\{t,\ldots, t+N-1\}$.
We choose L1 adaptation for comparison since it has been successfully applied to quadrotors~\cite{wu2025l1quad} and quadrupeds~\cite{sombolestan2024adaptive} recently for online adaptation.

\begin{figure*}[t]
    \centering
    \subfigure[Flat terrain with $\boldsymbol{f}_{u} = 12\boldsymbol{g}$.]{\includegraphics[width=0.49\textwidth]{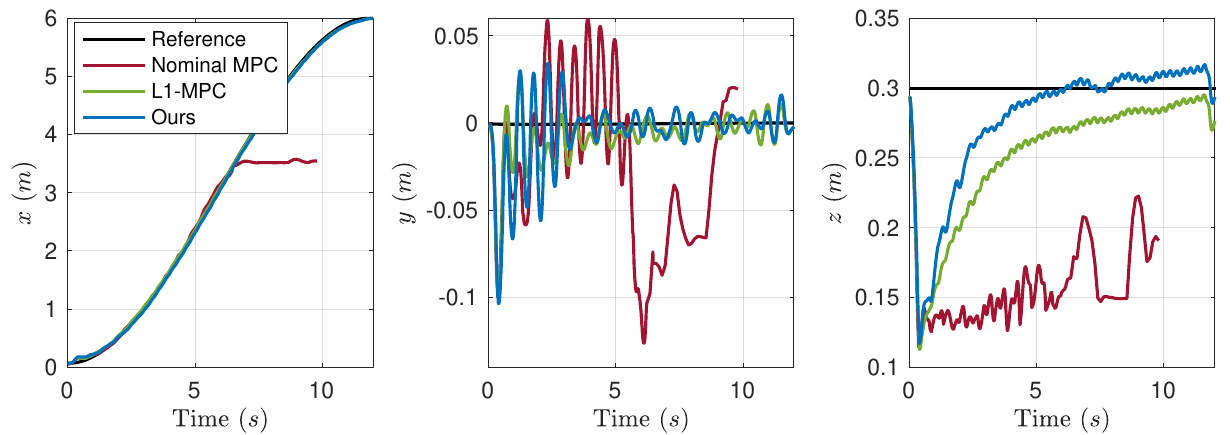}}\label{fig:sim-flat-traj}
    \subfigure[Slope terrain with $\boldsymbol{f}_{u} = 12\boldsymbol{g}$.]{\includegraphics[width=0.49\textwidth]{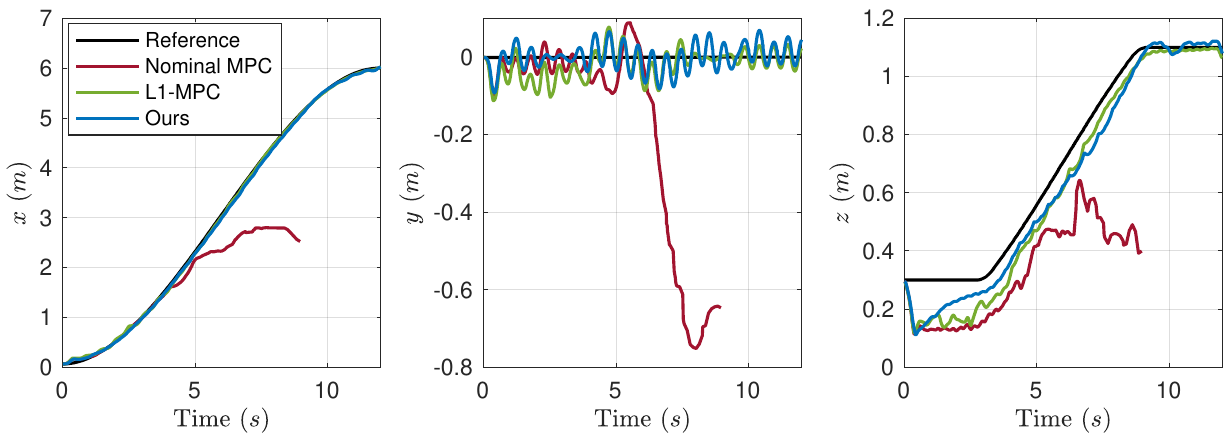}}\label{fig:sim-slope-traj}
    \subfigure[Rough terrain with \ensuremath{\boldsymbol{f}_{u} = 2 \left[ \|\boldsymbol{g}\|,\;0,\;\|\boldsymbol{g}\| \right]^\top}: Ours vs \NMPCf.]{\includegraphics[width=0.49\textwidth]{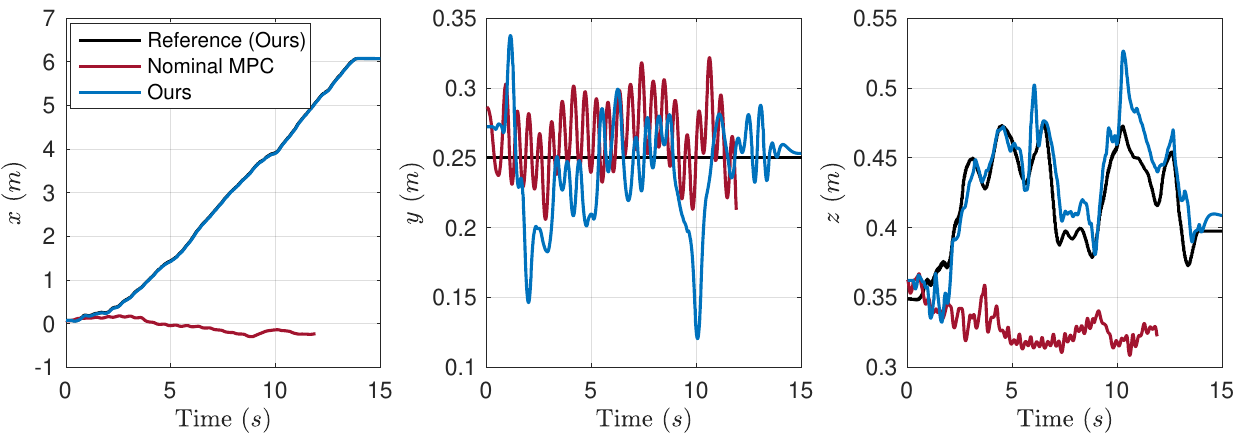}}\label{fig:sim-rough-xz-traj-ours}
    \subfigure[Rough terrain with \ensuremath{\boldsymbol{f}_{u} = 2 \left[ \|\boldsymbol{g}\|,\;0,\;\|\boldsymbol{g}\| \right]^\top}: \LMPCf vs \NMPCf.]{\includegraphics[width=0.49\textwidth]{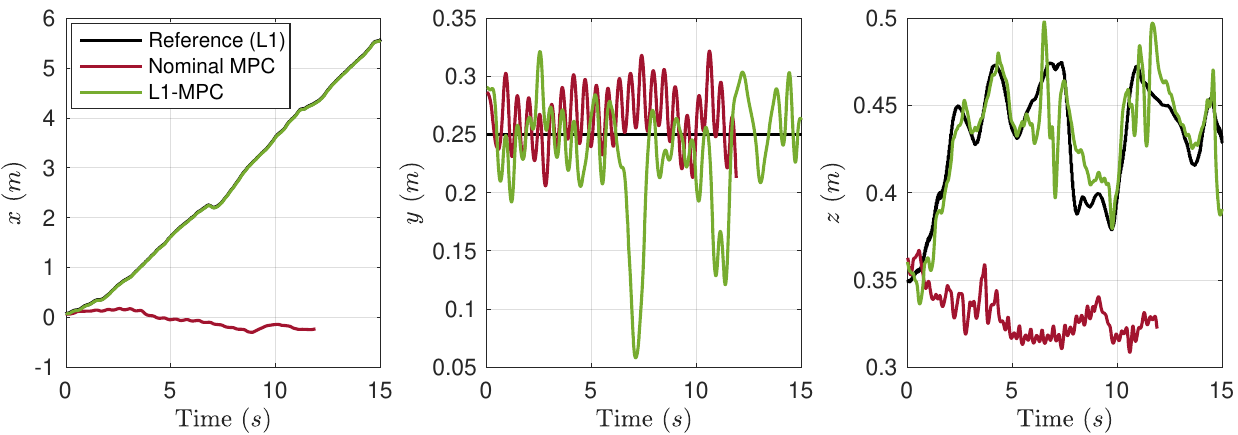}}\label{fig:sim-rough-xz-traj-l1}
    \caption{\textbf{Sample trajectories of the Gazebo Simulations in \Cref{subsec:sim-gazebo}.} Three scenarios in flat, slope, and rough terrains are provided. Our method enables the quadruped to reach the goal position while the nominal \MPC fails. 
    {We plot our method and \LMPC separately in the case of rough terrain since they move forward at different speeds due to $x$-direction forces and results in different reference trajectories in $x$ and $z$.}
    (a) and (b): The nominal \MPC fails due to the heavy load. {(c) and (d):  The nominal \MPC fails to move forward due to the $x$-component of $\boldsymbol{f}_{u}$. Our method exhibits faster response to $\boldsymbol{f}_{\boldsymbol{u}}$ than \LMPC. (a) and (b): faster tracking in $z$-direction, (c) and (d): faster tracking in $x$-direction despite the same velocity command, \ie ours reaches $x=6~m$ faster.}}
    \label{fig:sim-result-traj}
\end{figure*}

The results are given in \Cref{table:sim} and \Cref{fig:sim-result-traj}. 
In \Cref{table:sim},  all algorithms perform similarly when $\boldsymbol{f}_{u}=\boldsymbol{0}$, as the nominal model is sufficient to capture the quadruped dynamics. Across the scenarios when $\boldsymbol{f}_{u}$ is non-zero,  \Cref{alg:MPC} demonstrates significant improvement over the nominal \MPC in terms of overall tracking error. Specifically, \Cref{alg:MPC} achieves $67\%$ improvement in the case of slope terrain with  $\boldsymbol{f}_{u} = 8\boldsymbol{g}$.
Compared to \LMPC, \Cref{alg:MPC} achieves better tracking performance as terrain and external forces become complicated, demonstrating the benefit of learning a model instead of using a vector-value in \MPC. Specifically, \Cref{alg:MPC} achieves $21\%$ improvement in the case of rought terrain with  $\boldsymbol{f}_{u} = 2 \left[ \|\boldsymbol{g}\|,\;0,\;\|\boldsymbol{g}\| \right]^\top$. In the case of slope terrain with $\boldsymbol{f}_{u} = 12\boldsymbol{g}$, we observe that both algorithms perform similarly as the quadruped reaches its limit of handling uncertainty.

In addition, \Cref{alg:MPC} enables the quadruped to reach the goal position while the nominal \MPC fails, shown in \Cref{fig:sim-result-traj}. In flat and slope terrains with $\boldsymbol{f}_{u} = $, the nominal \MPC fails due to the heavy load. In rough terrain with $\boldsymbol{f}_{u} = 2 \left[ \|\boldsymbol{g}\|,\;0,\;\|\boldsymbol{g}\| \right]^\top$, the nominal \MPC fails to move forward due to the $x$-component of $\boldsymbol{f}_{u}$.
Compared to \LMPC, \Cref{alg:MPC} exhibits faster response to $\boldsymbol{f}_{\boldsymbol{u}}$, \eg $z$-direction tracking in flat and slope terrains~(\Cref{fig:sim-result-traj}~(a) \& (b)) and $x$-direction tracking in rough terrain~(\Cref{fig:sim-result-traj}~(c) \& (d)). Despite being given the same velocity command, our method enables the quadruped to walk faster in the $x$-direction.

\begin{figure}[t]
    \centering
    \includegraphics[width=0.49\textwidth]{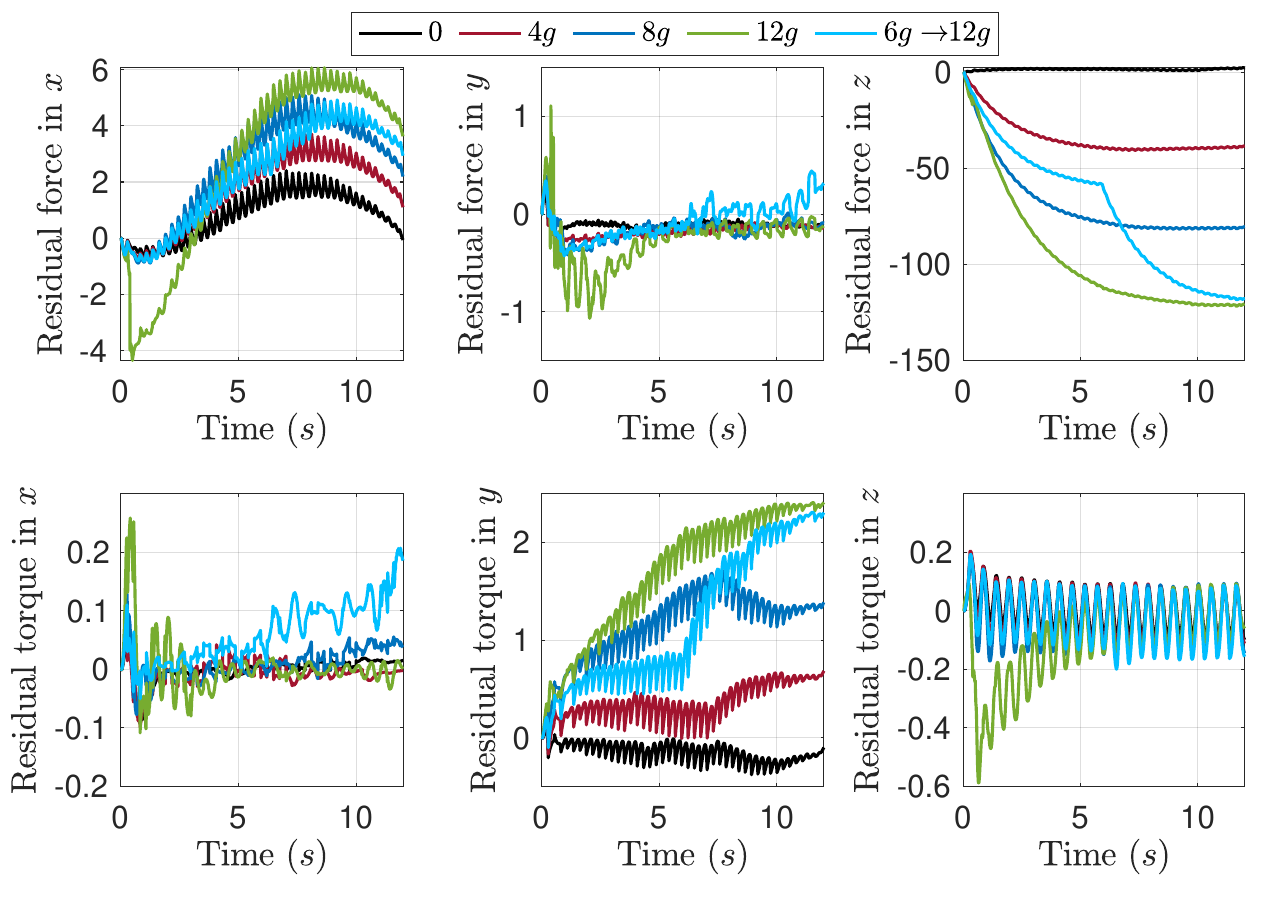}
    \caption{\textbf{Learned residual forces and torques $\hat{\boldsymbol{h}}\left(\boldsymbol{z}_{t};\hat{\boldsymbol{\alpha}}_t\right)$ over flat terrain with different $\boldsymbol{f}_u$ in \Cref{subsec:sim-gazebo}.} The main residual dynamics come from the force in $z$-direction, to which $\hat{\boldsymbol{h}}\left(\boldsymbol{z}_{t};\hat{\boldsymbol{\alpha}}_t\right)$ converges as the online learning module collects more data on-the-fly. This demonstrates that the online learning module is able to adapt to both constant and time-varying residual dynamics.}
    \label{fig:res}
\end{figure}

\Cref{fig:res} shows the learned residual forces and torques $\hat{\boldsymbol{h}}\left(\boldsymbol{z}_{t};\hat{\boldsymbol{\alpha}}_t\right)$ over flat terrain with different $\boldsymbol{f}_u$: (i) constant $\boldsymbol{f}_{u} = \boldsymbol{0}$, $4\boldsymbol{g}$, $8\boldsymbol{g}$, and $12\boldsymbol{g}$, and  (ii) time-varying $\boldsymbol{f}_{u}$ which switches from $6\boldsymbol{g}$ to $12\boldsymbol{g}$ when $x$-position reaches $3m$.
As expected, the main residual dynamics come from the force in $z$-direction, to which $\hat{\boldsymbol{h}}\left(\boldsymbol{z}_{t};\hat{\boldsymbol{\alpha}}_t\right)$ converges as the online learning module collects more data on-the-fly.
This demonstrates that the online learning module is able to adapt to both constant and time-varying residual dynamics.

\subsection{MuJoCo Simulations}\label{subsec:sim-mujoco}

\myParagraph{Simulation Setup} 
We evaluate \Cref{alg:MPC} on a \textit{sim2sim} setting. Specifically, we employ the controller used in \Cref{subsec:sim-gazebo} to control the Unitree Go2 Robot in the high-fidelity physics engine DeepMind MuJoCo \cite{todorov2012mujoco}.\footnote{\url{https://github.com/unitreerobotics/unitree_mujoco}} 
The control frequency and parameters are kept the same as in \Cref{subsec:sim-gazebo}.

\myParagraph{Benchmark Experiment Setup}
We consider flat terrains with constant and varying friction conditions, shown in \Cref{fig:sim-exp}~(first \& second). 
In the case of varying friction coefficients, the ground coefficients switch between $[0.5,  0.5,  0.01]$~(red rectangle) and $[0.05, 0.05, 0.001]$~(blue rectangle), which stands for sliding, torsional, and rolling frictions, respectively.
The quadruped is tasked to walk at $v_x = 0.5~m/s$ while maintaining its body height at $0.3~m$ above the ground. 
In both terrains, the quadruped carries either no payload or a payload weight at $0~kg$, $4~kg$, or $8~kg$. The $4~kg$ payload has inertia $[0.00234, 0.00304, 0.00414]~kg\cdot m^2$ and the $8~kg$ has $[0.00503, 0.00655, 0.00889]~kg\cdot m^2$.
The payload and varying ground friction coefficients create time-varying disturbances in both forces and torques: (i) payloads with mass and inertia that create state-dependent forces and torques, and (ii) varying ground friction coefficients that affect the ground reaction forces, therefore creating both force and torque disturbances.
Note that under the scenario of no payload, the nominal model in \MPC used in \Cref{subsec:sim-gazebo} has around $0.8~kg$ mismatch to the model simulated in MuJoCo, due to different weights of knee motors.
We use the tracking error as the performance metric.

\begin{figure*}[htbp]
    \centering
    \subfigure[No payload.]{\includegraphics[width=0.32\textwidth]{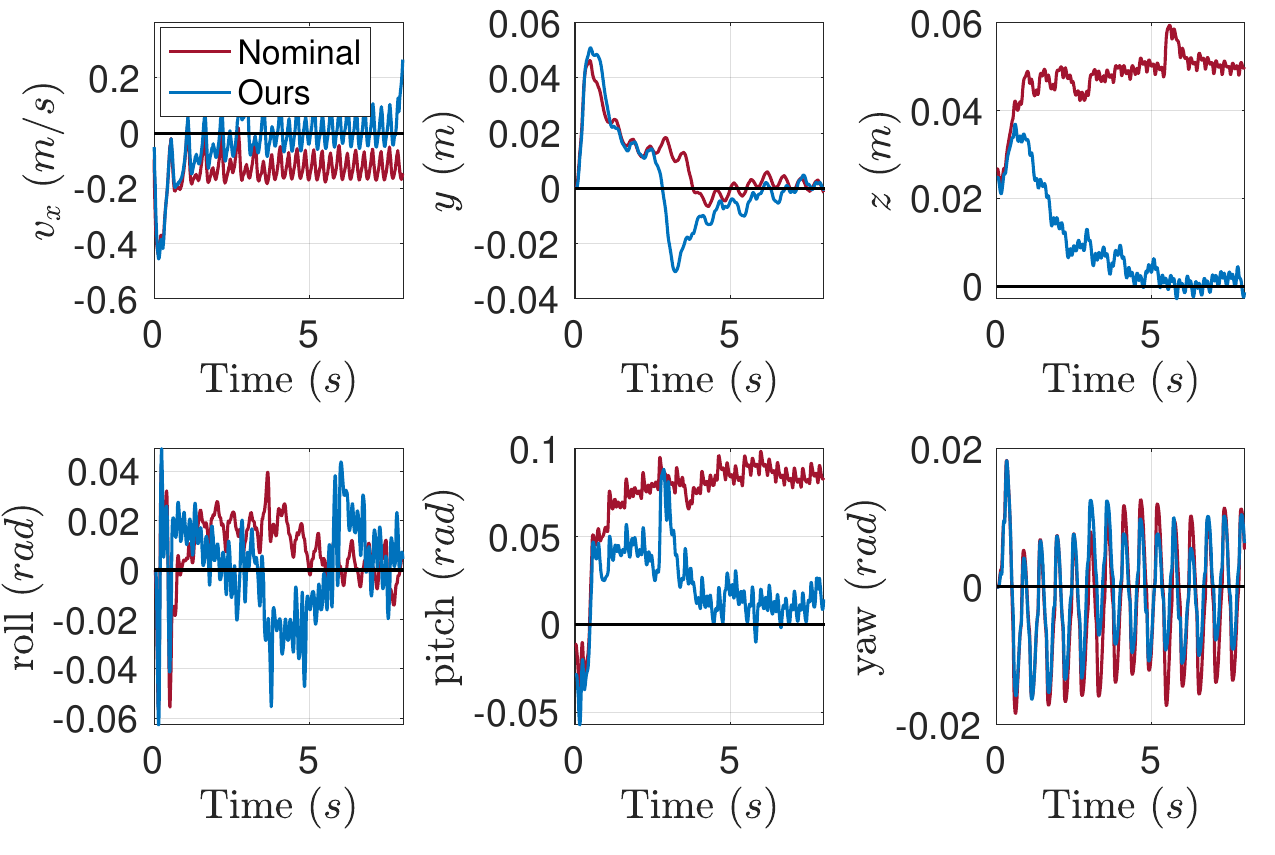}}\label{fig:mujoco-flat-0}
    \subfigure[$4~kg$ payload.]{\includegraphics[width=0.32\textwidth]{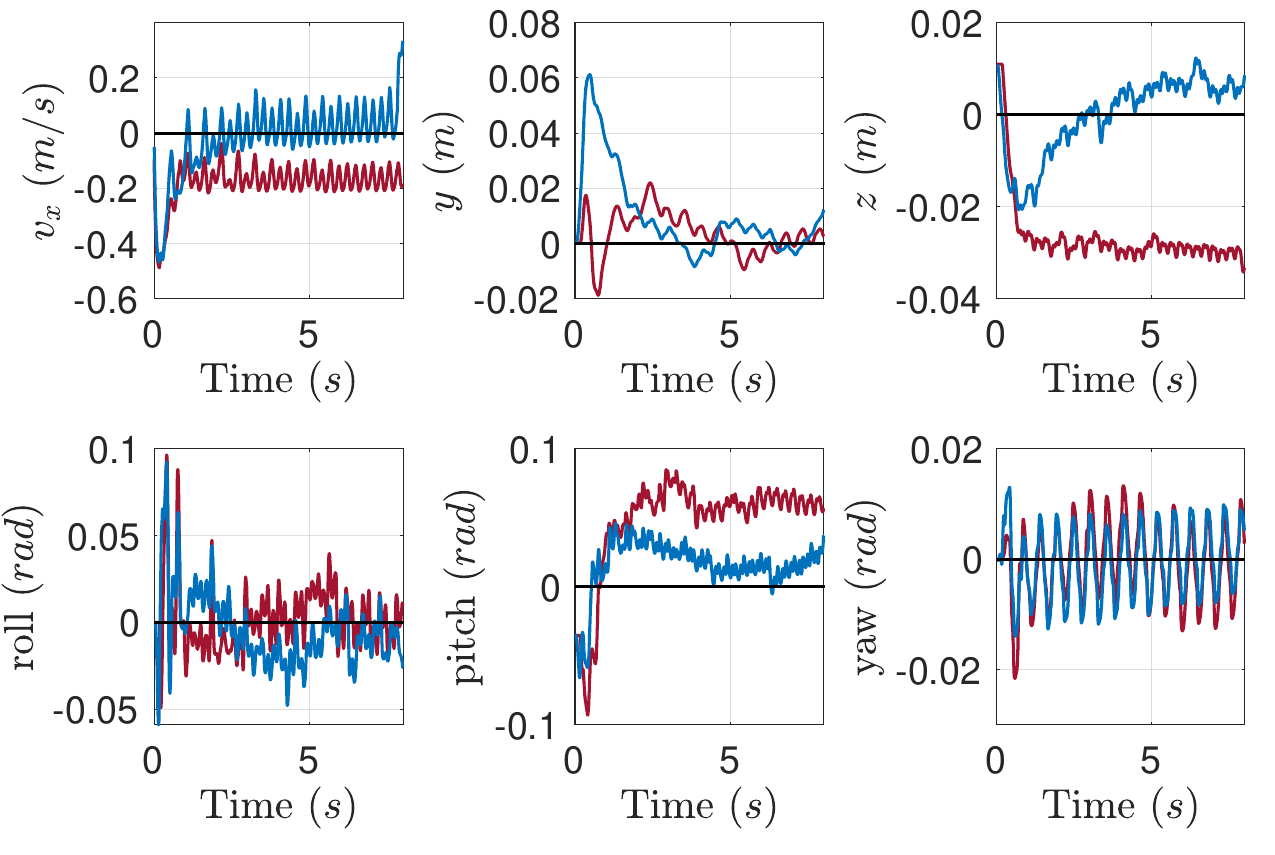}}\label{fig:mujoco-flat-4}
    \subfigure[$8~kg$ payload.]{\includegraphics[width=0.32\textwidth]{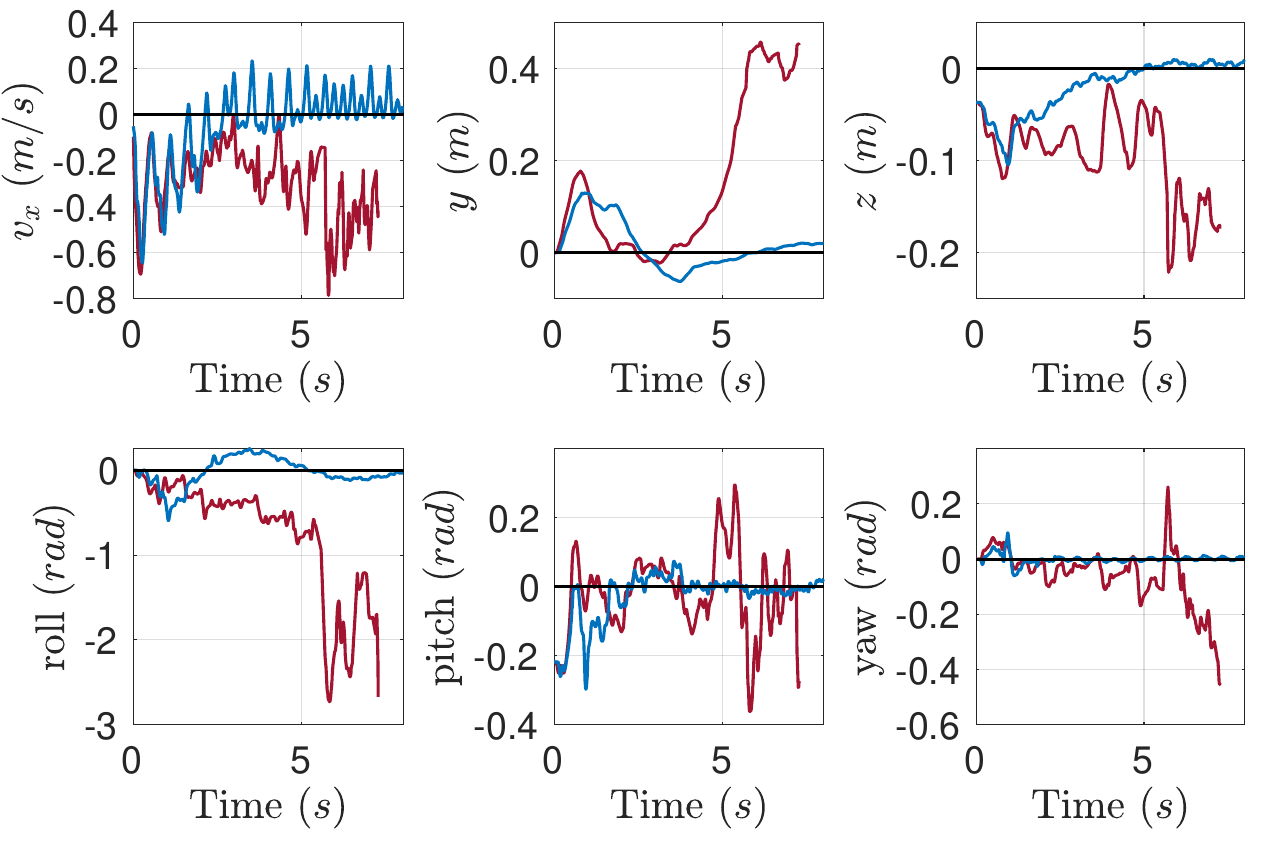}}\label{fig:mujoco-flat-8}
    \caption{{\textbf{Sample trajectories of the MuJoCo Experiments with constant friction coefficient in \Cref{subsec:sim-mujoco}.}  Scenarios with no payload, $4~kg$, and $8~kg$ payload are provided. Our method enables the quadruped to track the $0.5~m/s$ velocity command while maintaining the $0.3~m$ body height under different payload conditions. While the \NMPC has larger tracking errors in $z$ and $v_x$, and fails under $8~kg$ payload.}}
    \label{fig:sim-mujoco-result-flat}
\end{figure*}

\begin{figure*}[htbp]
    \centering
    \subfigure[No payload.]{\includegraphics[width=0.32\textwidth]{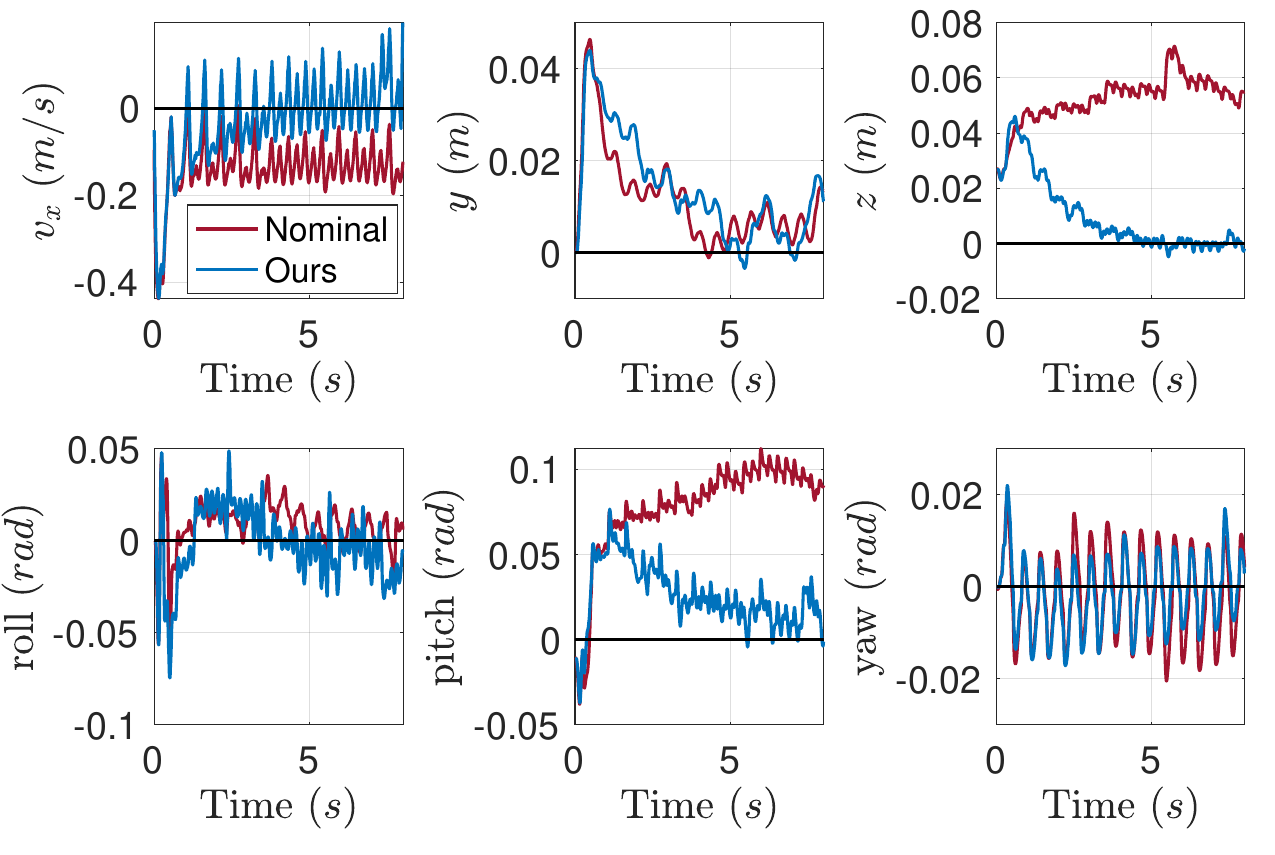}}\label{fig:mujoco-fric-0}
    \subfigure[$4~kg$ payload.]{\includegraphics[width=0.32\textwidth]{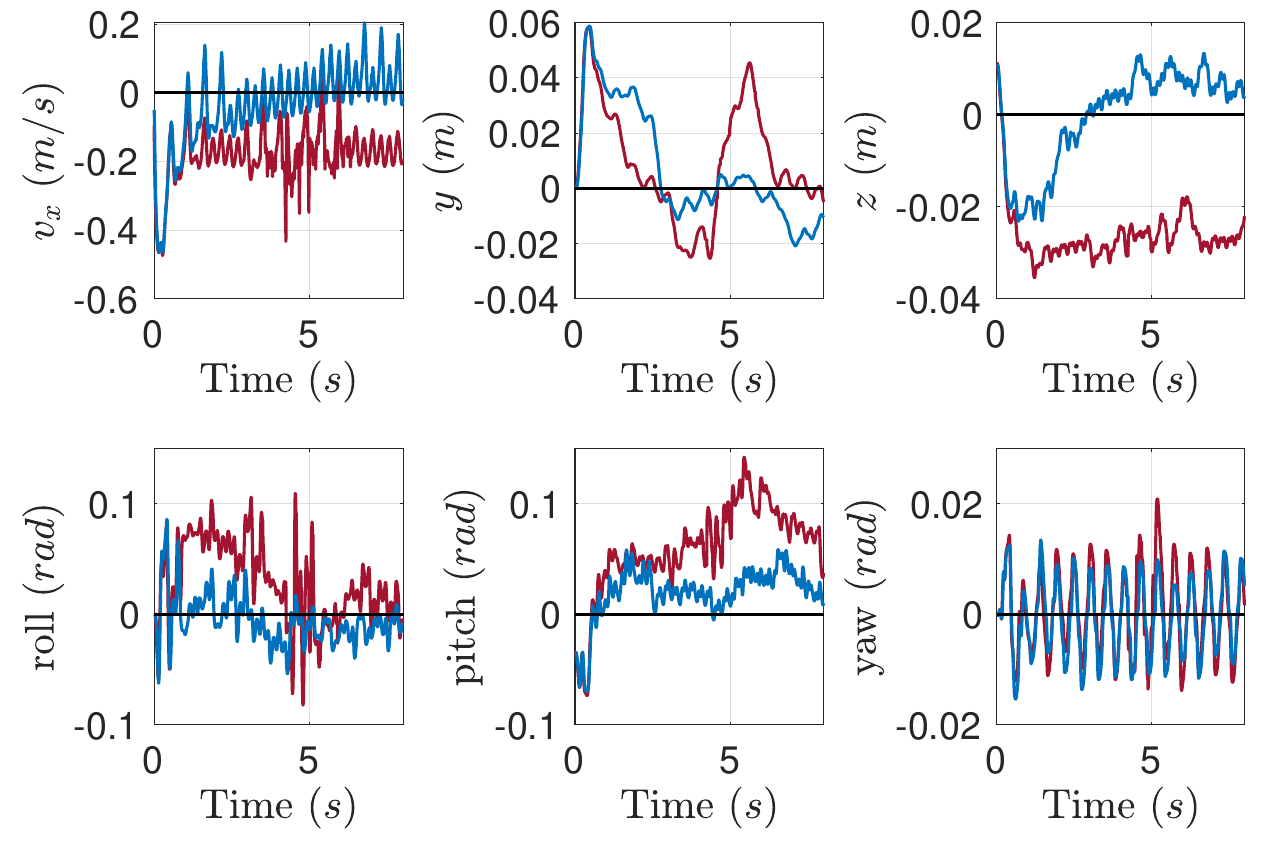}}\label{fig:mujoco-fric-4}
    \subfigure[$8~kg$ payload.]{\includegraphics[width=0.32\textwidth]{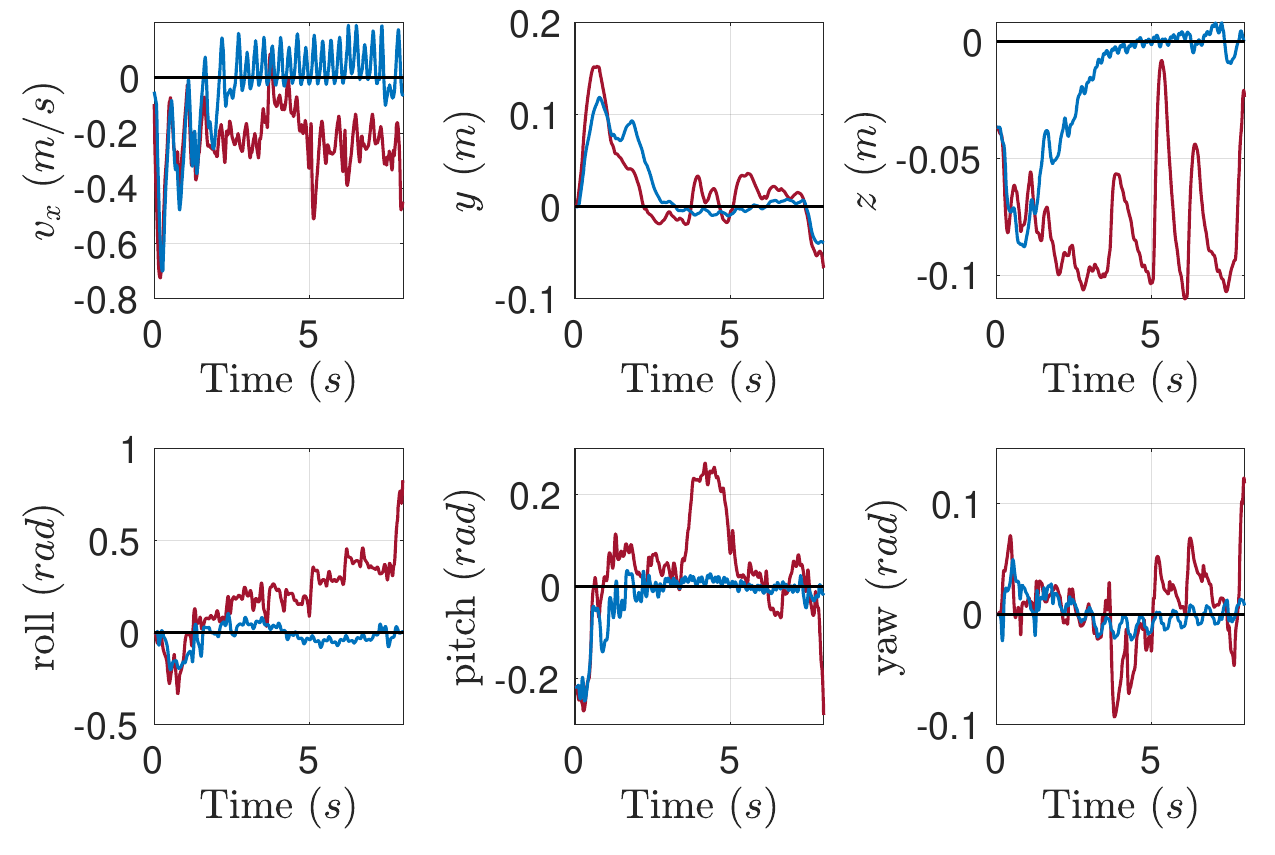}}\label{fig:mujoco-fric-8}
    \caption{{\textbf{Sample trajectories of the MuJoCo Experiments with varying friction coefficients in \Cref{subsec:sim-mujoco}.} Scenarios with no payload, $4~kg$, and $8~kg$ payload are provided.  Our method enables the quadruped to track the $0.5~m/s$ velocity command while maintaining the $0.3~m$ body height under different payload conditions. While the \NMPC has larger tracking errors in $z$ and $v_x$, and fails under $8~kg$ payload.}}
    \label{fig:sim-mujoco-result-fric}
\end{figure*}

\myParagraph{Results} 
The results are given in \Cref{fig:sim-mujoco-result-flat}~(constant friction) and \Cref{fig:sim-mujoco-result-fric}~(varying friction). 
In both terrains, our method enables the quadruped to track the $0.5~m/s$ velocity command while maintaining the $0.3~m$ body height under different payload conditions. While the \NMPC has larger tracking errors in $z$ and $v_x$, and fails under $8~kg$ payload.

\subsection{Failure Cases}\label{subsec:sim-fail}
Despite the online learning module for adapting to the residual dynamics, the method with L1 or random Fourier features may still fail under extreme conditions, such as over $4~g$ in rough terrain in the Gazebo simulations and $8~kg$ in changing friction coefficients in MuJoCo simulations. 
In those cases, the difficulty of stable walking or learning residual comes from sudden changes in foothold positions or unexpected contact.

\subsection{Discussion on Real-Time Computation}\label{subsec:sim-comp}
All the experiments are run on a computer with i7-13700k and 32 GB RAM.
Per \Cref{alg:MPC}, updates of $\boldsymbol{\alpha}$ are carried out at every control cycle; that is, the gradient descent step is executed every time before the \MPC is solved. 
In our implementation, this requires calling the nominal model and using the odometry information to obtain the "ground truth" disturbance, then using the learned residual dynamics to obtain the estimation loss. The gradient for parameter update can be obtained analytically from $l_t\left(\hat{\boldsymbol{\alpha}}_t\right) \triangleq \left\| \boldsymbol{h}\left(\boldsymbol{z}_{t}\right) -  \frac{1}{M} \sum_{i=1}^{M} \boldsymbol{\Phi}\left(\boldsymbol{z}_{t}, \boldsymbol{\theta}_i \right) \hat{\boldsymbol{\alpha}}_{i,t} \right\|^2$ due to the linearity of $\hat{\boldsymbol{h}}$ in $\hat{\boldsymbol{\alpha}}$.
Therefore, the online update process is a lightweight add-on to the original \MPC loop.
On the other hand, the residual dynamics $\hat{\boldsymbol{h}}$ add complexity to the nominal model, and the \MPC solve time will increase as $M$ increases. In our experiments, we use $M=50$ so that the control frequency remains $200~Hz$.
The complexity of the nominal dynamics also affects the real-time computation. Residual dynamics in joint space can be captured by incorporating joint dynamics into nominal dynamics, but this increases the state dimension and introduces additional computational burden.

\section{Conclusion} \label{sec:con}

We provided \Cref{alg:MPC} for \textit{Adaptive Legged Locomotion via Online Learning and Model Predictive Control}~(\Cref{prob:control}). 
The algorithm is composed of two interacting components: \MPC and online learning of residual dynamics.
The algorithm uses random Fourier features to approximate the residual dynamics in reproducing kernel Hilbert spaces. Then, it employs \MPC based on the current learned model of the residual dynamics.
The model of the residual dynamics is updated online in a self-supervised manner using least squares based on the data collected while controlling the quadruped.
\Cref{alg:MPC} guarantees no dynamic regret against an optimal clairvoyant (non-causal) policy that knows the residual dynamics a priori~(\Cref{theorem:regret_OLMPC}). 

The proposed \Cref{alg:MPC} is validated in the simulation environment with high-fidelity physics engines. Our simulations include quadruped aiming to track a reference trajectory despite {constant} uncertainty up to $12\boldsymbol{g}$ in flat, slope, and rough terrains. 
The algorithm (i) achieves up to $67\%$ improvement of tracking performance over the nominal \MPC and $21\%$ improvement over \LMPC, and (ii) succeeds even when nominal \MPC fails.
We also validate \Cref{alg:MPC} under time-varying uncertainty in flat terrains of different coefficients with up to $8~kg$ payload, showing the algorithm achieves significantly better tracking performance than \NMPC.



\bibliographystyle{IEEEtran}



\end{document}